\documentclass[journal]{IEEEtran}
\usepackage{cite}
\usepackage{hyperref}

\ifCLASSINFOpdf
  \usepackage[pdftex]{graphicx}
\else
\fi
\usepackage{amsmath}
\usepackage{amssymb}

\usepackage[caption=false,font=footnotesize]{subfig}
\usepackage{changes}
\usepackage{color}
\definecolor{r}{RGB}{192, 0, 0}
\usepackage{multirow}

\begin{document}
%
\title{MRDet: A Multi-Head Network for Accurate Rotated Object Detection in Aerial Images}
%
%
%

\author{Ran Qin, Qingjie Liu, 
	Guangshuai Gao, Di Huang, 
	and Yunhong Wang, 


}

%
%

\markboth{IEEE TRANSACTIONS ON GEOSCIENCE AND REMOTE SENSING, ~Vol.~14, No.~8, MARCH~2021}
{Shell \MakeLowercase{\textit{et al.}}: Bare Demo of IEEEtran.cls for IEEE Journals}
%



\maketitle

\begin{abstract}
Objects in aerial images usually have arbitrary orientations and are densely located over the ground, making them extremely challenge to be detected. Many of recent developed methods attempt to solve these issues by estimating an extra orientation parameter and placing dense anchors, which will result in high model complexity and computational costs. In this paper, we propose an arbitrary-oriented region proposal network (AO-RPN) to generate oriented proposals transformed from horizontal anchors. The AO-RPN is very efficient with only a few amounts of parameters increase than the original RPN. Furthermore, to obtain accurate bounding boxes, we decouple detection task into multiple subtasks and propose a multi-head network to accomplish them. Each head is specially designed to learn the features optimal for the corresponding task, which allows our network to detect objects accurately. We name it MRDet short for Multi-head Rotated object Detector. We evaluate the performance of the proposed MRDet on two challenging benchmarks, i.e., DOTA and HRSC2016, and compare it with several state-of-the-art methods. Our method achieves very promising results which clearly demonstrate its effectiveness.  Code has been available at \url{https://github.com/qinr/MRDet}.

\end{abstract}


%
\IEEEpeerreviewmaketitle

\section{Introduction}
%
%
%
%
\IEEEPARstart{O}{bject} detection is one of the fundamental tasks in computer vision, whose aim is to estimate categories of object instances in scenes and mark their locations, simultaneously. With the development of deep convolutional neural networks (DCNNs), object detection has achieved great advances in natural scenes, inspiring researchers in the remote sensing (RS) community to solve the intractable ground object recognition problem with deep learning techniques. 

Captured by optical sensors in a bird's-eye perspective from a great distance, objects in remote sensing images have distinctly different characteristics from those captured by consumer cameras. They may be placed on the ground with arbitrary orientations, thus making them hard to be covered by horizontal bounding boxes that widely used in general object detection frameworks. In addition, many RS objects such as ships and vehicles are with small sizes and usually densely packed, which will put a heavy burden on detection models. 

To conquer these issues, a number of approaches \cite{DBLP:journals/lgrs/LiuWWY16, DBLP:journals/tgrs/ZhangLZ19,DBLP:conf/iccv/YangYY0ZGSF19,DBLP:conf/icip/LiuHWY17,DBLP:conf/accv/AzimiVB0R18,fu2020rotation,DBLP:journals/corr/abs-1911-09358,DBLP:journals/lgrs/ZhangGZY18} have been developed. Many of them follow so-called general object detection frameworks such as Faster RCNN\cite{DBLP:conf/nips/RenHGS15} and adapt themselves to capture intrinsic features of objects in aerial images. Faster RCNN\cite{DBLP:conf/nips/RenHGS15} consists of two stages: a Region Proposal Network (RPN) that generates Horizontal Region of Interests (HRoIs) and a detection head that predicts locations and categories of regions. In natural images, objects are annotated with horizontal bounding boxes. To represent objects with arbitrary orientations accurately, a typical solution is to rotate the predicted bounding boxes to fit the object orientations with certain angles estimated from the object features carved by a set of anchors. Because predicting angles is a highly nonlinear task, it is difficult to obtain precise angles from the horizontal proposals. To alleviate this issue, some works \cite{DBLP:conf/icip/LiuHWY17, DBLP:journals/lgrs/ZhangGZY18, DBLP:conf/accv/AzimiVB0R18} design rotated anchors and regress them to Rotated Region of Interests (RRoIs) in the first stage.This will  lead to large number of anchors and thus suffer from high computations in both the training and testing stages.  

Note that rotated bounding boxes can be derived from the horizontal ones through some transformations with respect to rotation, scale, displacement, and so on.
RoI Transformer\cite{DBLP:conf/cvpr/DingXLXL19} acquires RRoIs by a lightweight spatial transformer next to the RPN stage. However, it introduces extra fully connected layers \((fcs)\) with many parameters. The model still needs flexible region proposal network design. To balance the accuracy and efficiency of detection, we improve RPN to generate arbitrary-oriented proposals with negligible parameter increase.

As aforementioned, object detection is comprised of two subtasks, a classification task and a localization task. The classification task should identify an object's category correctly regardless of its location, scale and orientation. And the localization task predicts a tight bounding box relevant to an instance's geometric configuration. Therefore, features suitable for classification and location are not the same. Bounding boxes with high classification confidences may have low Intersection over Unions (IoUs) with the matched ground truths\cite{DBLP:conf/eccv/JiangLMXJ18,DBLP:journals/corr/abs-1908-05641,DBLP:conf/cvpr/SongLW20,wu2019double}. IoU-Net\cite{DBLP:conf/eccv/JiangLMXJ18} designs a branch estimating IoUs and chooses boxes performing well on category identification as well as location in the post-processing period. However, IoU-Net still obtains the classification scores and locations with a shared head, which does not solve the essential issue on extracting respective features for different tasks. Song et al. \cite{DBLP:conf/cvpr/SongLW20} and Wu et al. \cite{wu2019double} seperate the shared head into two sibling heads for classification and localization, respectively. Double Head RCNN\cite{wu2019double} focuses on the network's architecture, which shows that fully connected head has more correlation between classification scores and IoUs, convolutional head is more suitable for localization task. TSD\cite{DBLP:conf/cvpr/SongLW20} aims to spatially disentangle sampling points of classification and localization. 

Inspired by these methods, we propose a multi-head network to predict classification, location, size and orientation of object instances, then integrate results from all the heads to obtain the final rotated bounding boxes and the class confidence scores. We name it MRDet short for Multi-head Rotated object Detector. MRDet is a two-stage approach following the paradigm of Faster RCNN \cite{DBLP:conf/nips/RenHGS15}. It consists of two modules, Arbitrary-Oriented Region Proposal Network (AO-RPN) and Multi-Head Network (MH-Net). In the first stage, AO-RPN generates inclined proposals by rotating the horizontal proposals using learnt transformations. In contrast to previous Rotated RPNs, our AO-RPN is efficient since it has the same number of anchors to horizontal proposals with a slight cost of a few amounts of parameters increase. In the final stage, MH-Net decouples the detection task into category classification, object location, scale estimation, and orientation prediction subtasks and realizes them with four sibling heads. To summarize, the main contributions of this paper are as follows:
\begin{itemize}
  \item [\(\bullet\)] 
  We design a novel Arbitrary-Oriented Region Proposal Network (AO-RPN) to generate HRoIs and RRoIs, simultaneously. 
  The network is efficient with only slightly computation increase than the original RPN. 
  \item [\(\bullet\)] 
  We propose a Multi-Head Network (MH-Net) to predict category scores, locations, scales and orientations of the objects, respectively. It can achieve more accurate detection performance by decomposing detection into four easier tasks and learning task-specific features. 
  \item [\(\bullet\)]
  Our method achieves state-of-the-art performance on two challenging aerial object detection benchmarks, i.e., DOTA\cite{DBLP:conf/cvpr/XiaBDZBLDPZ18} and HRSC2016\cite{DBLP:conf/icpram/LiuYWY17}, which clearly demonstrates its superiority. 
\end{itemize}

The remainder of this paper is organized as follows. Section \ref{related works} gives a brief review related to this work. Section \ref{our method} presents our proposed approach in detail. In Section \ref{experiments}, we conduct extensive experiments on two challenging aerial object detection datasets to validate the effectiveness of our method. Finally, we conclude the paper in Section \ref{conclusion}.

 




\section{Related Work}\label{related works}
\subsection{Generic Object Detection}
Object detection aims to localize specific object instances in images and mark them with bounding boxes. With the advancement of the deep learning techniques, object detection has achieved great progress thanks to the powerful representative ability of deep convolutional neural networks. According to the detection pipeline, most of the existing object detectors can be divided into two types: two-stage methods and one-stage methods. Two-stage detectors first generate a set of category-agnostic region of interests (RoIs) that potentially contain objects. This is achieved by region proposal networks (RPNs). Then in the second stage, head networks perform detection using a shared network for category prediction and location estimation. The most representative two-stage detectors are the pioneering RCNN family\cite{DBLP:conf/cvpr/GirshickDDM14, DBLP:conf/iccv/Girshick15, DBLP:conf/nips/RenHGS15}. To deal with scale variations, Feature Pyramid Network (FPN)\cite{DBLP:conf/cvpr/LinDGHHB17} takes advantage of the pyramid shape of convolution features and combines them in various resolutions to construct a feature pyramid with rich semantic information to recognize objects at different scales. Subsequently, \cite{DBLP:journals/corr/abs-1711-07264,DBLP:conf/nips/DaiLHS16,DBLP:conf/cvpr/CaiV18} are proposed to improve the computational speed and obtain higher detection accuracy. 

In contrast to two-stage detectors, one-stage methods get rid of the complex regional proposal stage (i.e., RPN) and predict the object instance categories and their locations directly from densely pre-designed candidate boxes. One-stage detectors are popularized by YOLO\cite{DBLP:conf/cvpr/RedmonDGF16}, SSD\cite{DBLP:conf/eccv/LiuAESRFB16} and RetinaNet\cite{DBLP:conf/iccv/LinGGHD17}. The main advantage of one-stage detectors is their high computational efficiency. However, the detection accuracy of one-stage detectors usually fall behind that of two-stage detectors, mainly because of the class imbalance problem. This gap was reduced by the Focal Loss solution \cite{DBLP:conf/iccv/LinGGHD17} and many other followers, such as \cite{DBLP:conf/aaai/Li0W19, DBLP:conf/cvpr/CaoCLL20}. Even so, when facing challenging scenarios such as small and densely packed objects, one-stage detectors are still unsatisfactory, and two-stage detectors are preferred solutions.

Recently, a new family of anchor-free detectors have arisen and gained increasing attention. The aforementioned detectors such as Faster RCNN\cite{DBLP:conf/nips/RenHGS15}, FPN \cite{DBLP:conf/cvpr/LinDGHHB17}, SSD \cite{DBLP:conf/eccv/LiuAESRFB16}, and RetinaNet \cite{DBLP:conf/iccv/LinGGHD17} place pre-defined anchor boxes densely over the feature maps and use them as references for bounding box regression and region candidates for classification prediction. Anchor-free methods believe that anchors are obstacle to further boosting detection performance and generalization ability \cite{DBLP:conf/nips/YangZLZS18, DBLP:journals/tip/KongSLJLS20}, and suggest learning the probabilities and bounding box coordinates of objects without anchor references. For instance, CornerNet\cite{DBLP:conf/eccv/LawD18} predicts the top-left and bottom-right corners, and groups them into bounding boxes. As corners are always located outside the objects, CenterNet\cite{DBLP:conf/iccv/DuanBXQH019} further adds geometric center points to obtain features inside objects which can decrease the matching error of two corners. ExtremeNet\cite{DBLP:conf/cvpr/ZhouZK19} attempts to predict key-points on geometric boundaries of instances. Although general object detection approaches have achieved great success in natural scenes, they perform poorly on arbitrary-oriented objects in aerial images.

\begin{figure*}[!t]
  \centering
  \includegraphics[width=18cm]{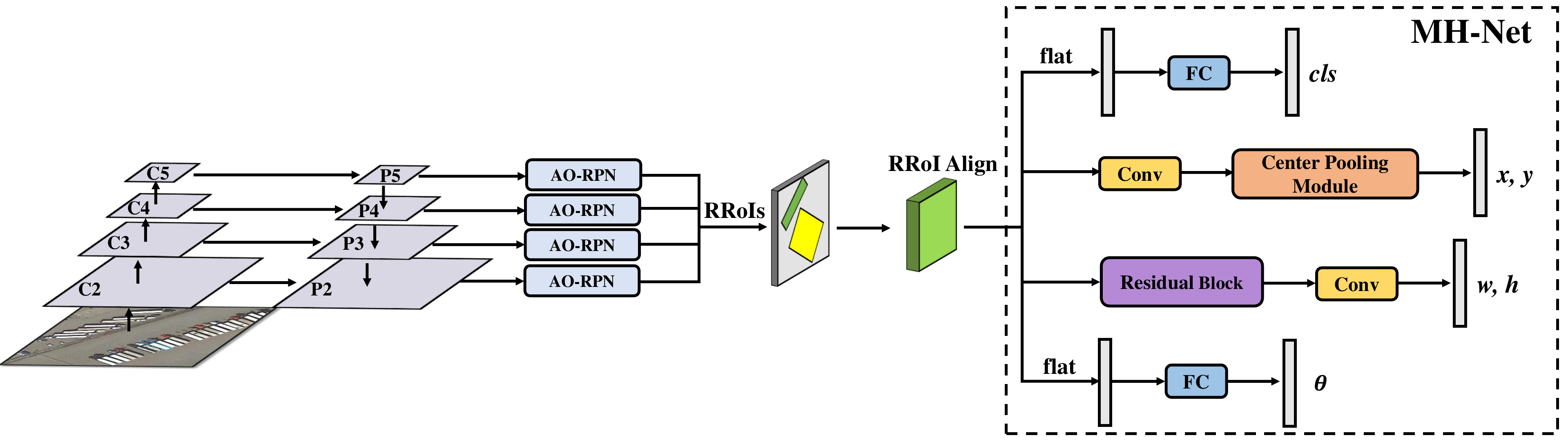}
  \caption{The architecture of MRDet. We use the fashionable FPN \cite{DBLP:conf/cvpr/LinDGHHB17} as the base model to build our method. Then we use Arbitrary-Oriented Region Proposal Network (AO-RPN) on each layer of FPN to generate rotated proposals with different scales. After RRoI Align, features inside RRoIs are sampled and fed in Multi-Head Network (MH-Net) for the final bounding boxes prediction. Particularly, the localization results \((x, y, w, h, \theta)\) are integrated from three sibling heads with different structures.}
  \label{the architecture of MR2Det}
  \end{figure*}

\subsection{Object Detection in Aerial Images}
The significant progress in general object detection has provided rich resources for developing aerial object detection models. Many works follow general object detection frameworks and adapt them to the aerial image domain. Early works \cite{DBLP:journals/tgrs/ChengZH16, DBLP:journals/tgrs/LongGXL17, DBLP:journals/lgrs/XiaoGLLWL17, DBLP:conf/cvpr/ChengZH16} adopt RCNN \cite{DBLP:conf/cvpr/GirshickDDM14} pipeline to detect objects with arbitrary orientations.
Cheng et al. \cite{DBLP:journals/tgrs/ChengZH16} design a rotation-invariant layer to enforce objects with and without rotation to have similar features. Xiao et al. \cite{DBLP:journals/lgrs/XiaoGLLWL17} fuse multi-scale features to include context information for airport detection in complex backgrounds. These methods have shown impressive performances in detecting aerial objects, however, they also inherit the inherent limitations of RCNN that the candidates feeding into the networks are generated by manually sliding window methods which is inefficiency and time-consuming. Later works \cite{DBLP:conf/icip/LiuHWY17, DBLP:conf/accv/AzimiVB0R18, fu2020rotation, DBLP:journals/lgrs/ZhangGZY18} embrace the milestone detector Faster RCNN \cite{DBLP:conf/nips/RenHGS15} and improve it with a rotated RPN\cite{DBLP:journals/tmm/MaSYWWZX18}, which can generate more accurate bounding boxes. However, extra anchors (i.e., rotated anchors) are introduced, resulting in lower computational efficiency. Inspired by \cite{DBLP:conf/nips/JaderbergSZK15}, Ding et al. \cite{DBLP:conf/cvpr/DingXLXL19} introduce a RoI Transformer to model geometry transformation of horizontal  RoIs. It shows a good trade-off between efficiency and detection accuracy.

In addition to two-stage detectors, some studies focus on adapting one-stage detection pipeline to develop real-time detectors. Sharing the idea with SSD\cite{DBLP:conf/eccv/LiuAESRFB16}, Tang et al. \cite{DBLP:journals/remotesensing/TangZDLZ17} add an extra parameter to predict rotation angles of bounding boxes. Dynamic refinement network (DRN)\cite{DBLP:conf/cvpr/PanRSDYGMX20}, built based on CenterNet\cite{DBLP:conf/iccv/DuanBXQH019}, predicts angles, sizes and location offsets from features convolved with controllable kernels, which intends to solve the misalignments between objects and receptive fields.  In addition to describing target locations as rotated rectangles, some methods consider a more flexible way by representing objects as quadrangles and predict them in a vertex-wise manner. These methods also achieve promising performance. For instance, DMPNet \cite{DBLP:conf/cvpr/LiuJ17} predicts the center of a quadrilateral and regresses the coordinates of its four vertexes by computing the relative offsets to the center. Similarly, RRD \cite{DBLP:conf/cvpr/LiaoZSXB18} and Gliding Vertex\cite{DBLP:journals/corr/abs-1911-09358} predict the offsets of four vertexes from the horizontal bounding box to the quadrilateral. As densely distributed objects in aerial images usually have regular shapes and similar orientations, rotated rectangles are more suitable than quadrilaterals. Therefore, in this paper, we further explore the spatial relationships between horizontal bounding boxes and rotated bounding boxes and propose a vertex-wise rotated proposal generation network based on affine transformations, such as scalings and rotations, instead of simple spatial displacements between vertexes. 

One key characteristic of objects in aerial images is that they are with small sizes and usually densely packed over grounds, and instances belonging to the same category, e.g., vehicles or ships always share almost identical appearances and similar orientations and scales. Under these scenarios, context and high-frequency co-occurrence could provide useful cues for recognizing objects. CAD-Net\cite{DBLP:journals/tgrs/ZhangLZ19} designs a global context network and a local context network to capture this information, simultaneously. Besides, Sun et al. \cite{SUN202150} propose a context refinement module which aggregates context in multiple layers to utilize both local and global context information. To detect small and cluttered objects, SCRDet\cite{DBLP:conf/iccv/YangYY0ZGSF19} introduces attention modules to suppress the noise and highlight the objects feature. SRAF-Net \cite{2020SRAF} combines context attention and deformable convolution to extract better features of objects with vague appearance and pay more attention to RoIs from the noisy environment.

\subsection{Classification v.s. Localization}
In order not to miss any objects, detectors tend to produce a large number of bounding boxes near the ground truth, and then to remove redundant boxes, a post processing step, i.e., Non-Maximum Suppression (NMS), is applied. An issue arises as a result of the mismatch between the highest scores and the best bounding boxes. Because the boxes with the highest scores may not be the best match to the object, the results may include boxes with high classification confidences however low IoUs with the corresponding ground truths, and vice versa, evidenced by the experiments in \cite{DBLP:conf/eccv/JiangLMXJ18,DBLP:journals/corr/abs-1908-05641}. This implies that the two tasks, i.e., classification and localization have opposite preferences with each other, inspiring researchers to rethink classification and localization for detection and design specific structures for them. Double Head RCNN\cite{wu2019double} finds that fully connected head is more suitable for the classification, as the classification score is more correlated to the IoU between a proposal and its corresponding ground truth. Convolution head regresses more accurate bounding boxes. As a result, the model disentangles the shared head into two structurally different branches: one fully-connected head for classification and one convolutional head for localization. TSD\cite{DBLP:conf/cvpr/SongLW20} focuses on disentangling spatial features for classification and localization. Each task seeks the optimal solutions for misalignment without interleaving to each other. In this work, we take a further step by dividing localization into three subtasks, i.e., center point localization, scale estimation, and orientation prediction, and exploring suitable architectures for each sibling task.

\section{Proposed Method}\label{our method}
In this section, we give details of our MRDet. The overall architecture is shown in Fig. \ref{the architecture of MR2Det}. In the first stage, AO-RPN generates rotated proposals efficiently without increasing the number of anchors, followed by a RRoI Align layer which extracts features of proposals. In the second stage, we devise a multi-branch head (MH-Net) to alleviate misalignments between features and subtasks. Detection is accomplished by producing classification scores, center locations, scales, and orientations of bounding boxes from corresponding branches.

\subsection{Arbitrary-Oriented Region Proposal Network (AO-RPN)}\label{Arbitrary-Oriented Region Proposal Network}
\begin{figure}[ht]
  \centering
  \includegraphics[width=7cm]{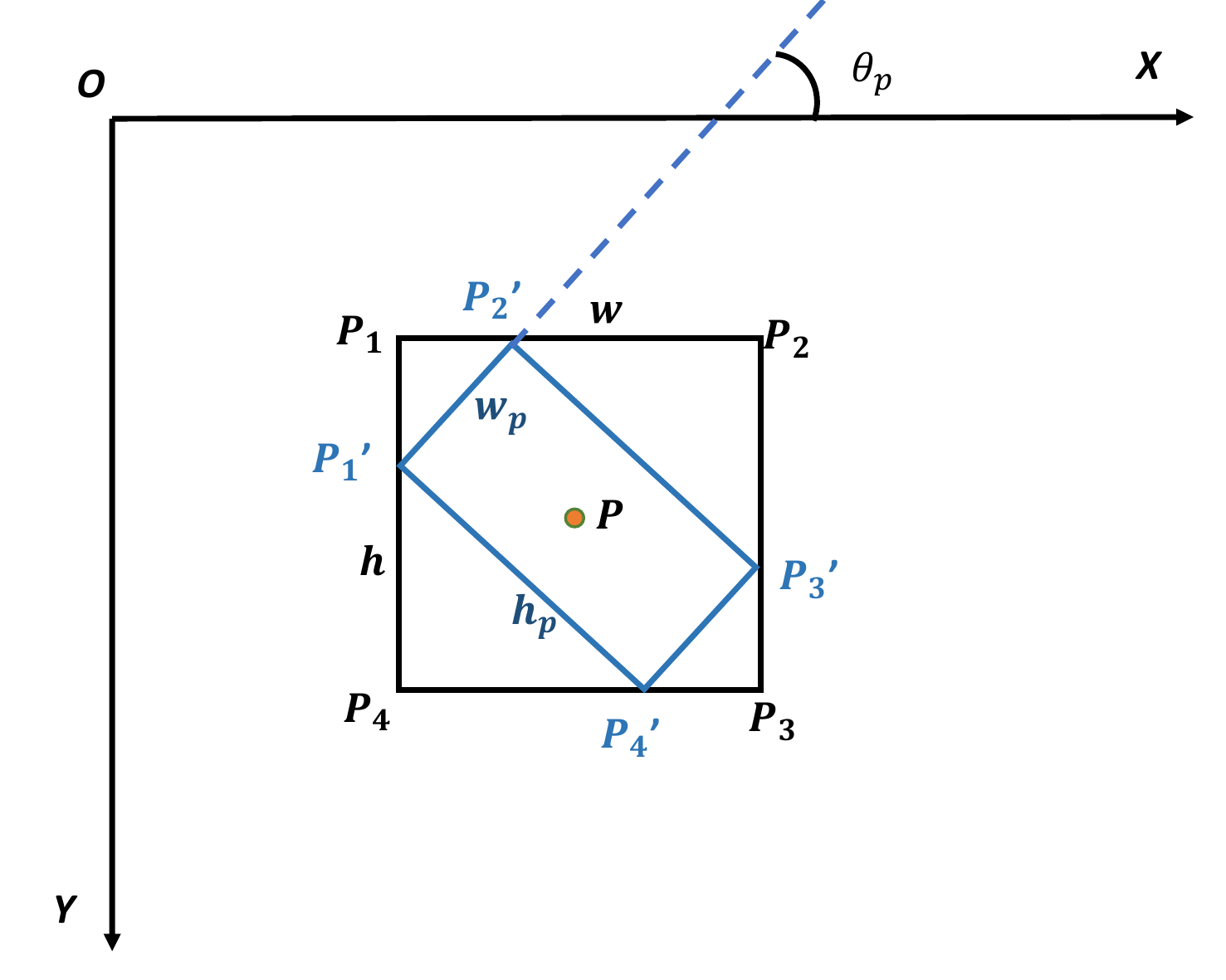}
  \caption{A sketch of horizontal enclosing rectangles and rotated bounding boxes. The horizontal box \((P_1, P_2, P_3, P_4)\) is enclosed by black lines and the rotated box \((P'_1, P'_2, P'_3, P'_4)\) by blue lines. The geometric center of the horizontal box and the rotated box are the same point \(P\).}
  \label{An example of horizontal enclosing rectangle and rotated bounding box}
\end{figure}

AO-RPN is a fully convolutional multi-task network, which is built on top of RPN \cite{DBLP:conf/nips/RenHGS15}, aiming to generate a set of category-agnostic  arbitrary-oriented proposals for subsequent usages. Horizontal proposals are first generated from pre-defined anchors as in RPN. Assume one proposal is denoted as (\(x, y, w, h\)), where (\(x, y\)) indicates the geometric center. The width \(w\) is set to the horizontal side and the height \(h\) is set to the vertical side. We then learn affine transformations to obtain oriented candidates from horizontal proposals. A rotated object region is represented as a 5-tuple (\(x_p, y_p, w_p, h_p, \theta_p\)). \(\theta_p\) represents the inclined orientation. Note that the geometric centers of the rotated proposals are the same to the horizontal ones as shown in Fig. \ref{An example of horizontal enclosing rectangle and rotated bounding box}, which indicates \((x, y) = (x_p, y_p)\). This leaves us only scaling and rotating  parameters to be learnt. Suppose \(P^{'}_i = (x^{'}_i, y^{'}_i)\) \((0\leq i<4)\) are vertexes of the rotated proposal and they can be calculated as follows:
\begin{equation}
M_\theta = \begin{pmatrix}\cos\theta_{p}&-\sin\theta_{p}\\\sin\theta_{p} & \cos\theta_{p}\end{pmatrix},
\label{rotation matrix}
\end{equation}

\begin{equation}
M_s = \begin{pmatrix}\frac{w_{p}}{w}&0\\0&\frac{h_{p}}{h}\end{pmatrix},
\label{scale matrix}
\end{equation}

\begin{equation}
\begin{pmatrix}x^{'}_{i}\\y^{'}_{i}\end{pmatrix}
= M_\theta * M_s * 
\begin{pmatrix}x_{i}-x\\y_{i}-y\end{pmatrix}
+
\begin{pmatrix}x\\y\end{pmatrix},
\label{equation for affine transformation}
\end{equation}
where \(P_i = (x_{i}, y_{i})\) is the vertex coordinate of a horizontal bounding box. \(M_\theta\), \(M_s\) denote rotating and scaling parameters of the affine transformation, respectively.  \(\theta_p\) is defined as the acute angle to the \(x\)-axis as shown in Fig. \ref{An example of horizontal enclosing rectangle and rotated bounding box},  \(\theta_p \epsilon (-\pi/2, \pi/2)\). Note that for horizontal bounding boxes, \(\theta = 0\). The order of the four vertexes is rearranged to minimize the angle as follows:
\begin{equation}
\label{theta define1}
\theta_p = \theta_i,
\end{equation}
\begin{equation}
\label{theta define2}
i = \arg\min_{0 \leq j < 4}\{|\theta_j - \theta|\},
\end{equation}
where \(\theta_p\) is the minimum angle rotated from a horizontal rectangle to its corresponding oriented rectangle.

\begin{figure}[ht]
  \centering
  \includegraphics[width=8cm]{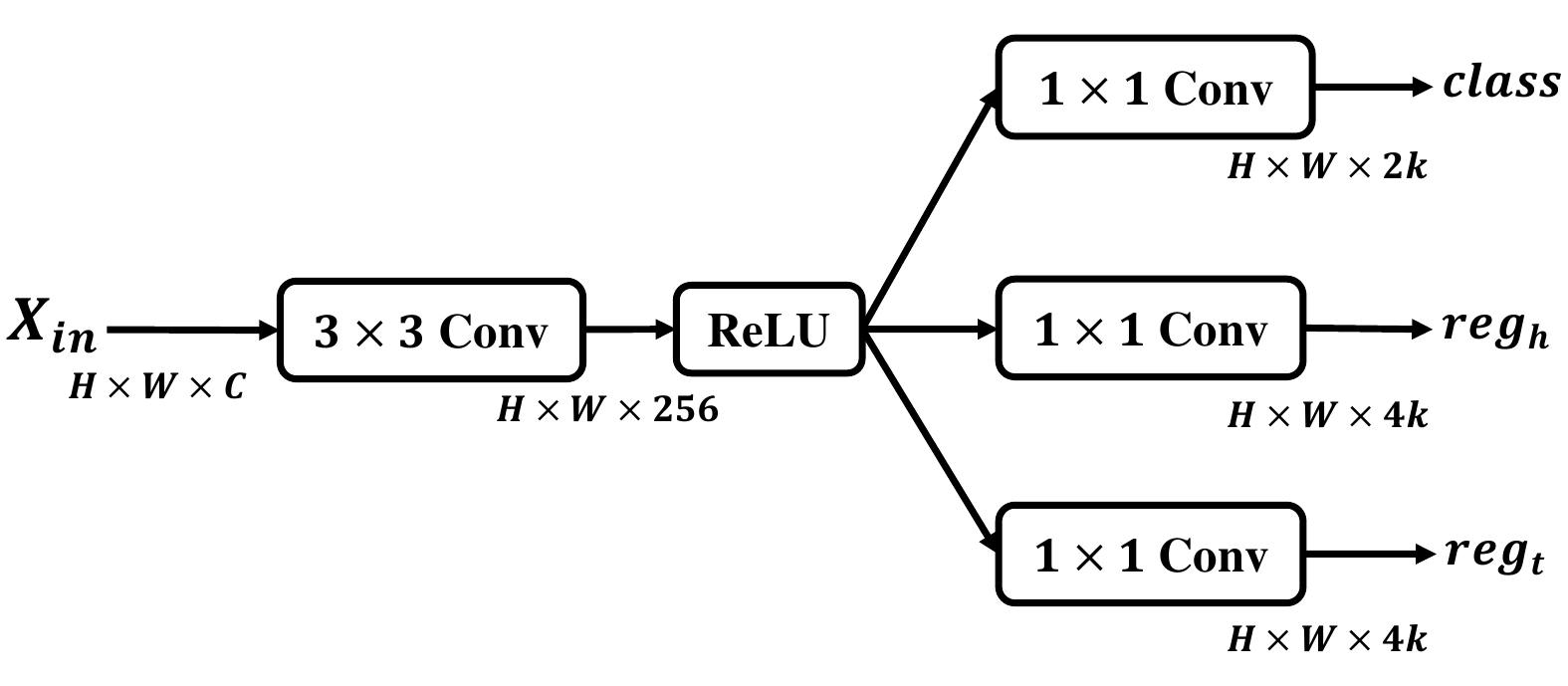}
  \caption{The structure of Arbitrary-Oriented Region Proposal Network (AO-RPN).}
  \label{Fig AO-RPN}
\end{figure}

The architecture of AO-RPN is illustrated in Fig. \ref{Fig AO-RPN}. A shared \(3\times 3\) convolution layer \((conv)\) follows the feature layer. Then three parallel branches constructed with \(1\times 1\) \(conv\)s are added to produce proposals, where \(class\) predicts binary confidence scores, \(reg_h\) outputs locations of the horizontal proposals, \(reg_t\) estimates the transformation parameters rotating the horizontal proposals defined by \(reg_h\) to the rotational ones. As can be seen, in contrast with RPN\cite{DBLP:conf/nips/RenHGS15}, AO-RPN only adds one more \(1\times 1\) \(conv\) branch to learn transformations for generating rotational proposals with very slight parameter increase. The shared features are fed into three sibling layers, and for each position in the feature map, there are \(k\) anchors. Therefore, the \(class\) layer has \(2k\) outputs estimating whether anchors belong to the background or the instance regions. The \(reg_h\) layer outputs \(4k\) points (\(u_x, u_y, u_h, u_w\)) for the minimum horizontal enclosing boxes corresponding to anchors. Besides, the \(reg_t\) layer generates \(4k\) transformation parameters (\(v_1, v_2, v_3, v_4\)) relative to horizontal proposals. The network outputs arbitrary-oriented proposals without increasing the number of anchors.

Following is the loss function to train AO-RPN,
\begin{equation}
  \begin{aligned}
  L(\{p_i\}, \{u_i\}, \{v_i\}) = {}&\frac{1}{N_{cls}}\sum_{i}L_{cls}(p_i, p_i^*) 
  {}\\& + \lambda_1\frac{1}{N_{reg}} \sum_{i}p_i^*L_{reg}(u_i, u_i^*)
  {}\\& + \lambda_2\frac{1}{N_{reg}} \sum_{i}p_i^*L_{reg}(v_i, v_i^*)
  \end{aligned}
\end{equation}
Here, \(i\) is the index for anchors. {\(p_i\)}, {\(u_i\)}, {\(v_i\)} indicate outputs of the \(class\) layer, \(reg_h\) layer and \(reg_t\) layer. \(p_i^*\) represents the classification label and means background when \(p_i^*=0\). \(u_i\), \(v_i\) and \(u_i^*\), \(v_i^*\) denote the predicted and ground truth of horizontal bounding box and oriented bounding box. \(\lambda_1\), \(\lambda_2\) are balance parameters. Empirically, we set \(\lambda_1=1\) and \(\lambda_2=1\). \(N_{cls}\) represents the number of sampled anchors and \(N_{reg}\) is assigned to the number of positive samples. We use cross entropy loss function for classification and smooth L1 loss for regression as follows:
\begin{equation}
L_{cls}(p_i, p_i^*) = -[p_i^*\log(p_i) + (1 - p_i^*)\log(1 - p_i)]
\end{equation}
\begin{equation}
L_{reg}(u_i, u_i^*) = Smooth_{L1}(u_i^* - u_i)
\end{equation}
\begin{equation}
Smooth_{L1}(x) =
\begin{cases}
0.5x^2,\quad\quad |x|< 1 \\
|x| - 0.5,\quad otherwise
\end{cases}
\end{equation}
The tuples \(u\), \(u^*\) encoding from horizontal proposals are calculated as:
\begin{equation}
\begin{aligned}
&u_x = \frac{x - x_a}{w_a}, u_y = \frac{y - y_a}{h_a},
\\&u_h = \log\frac{h}{h_a}, u_w = \log\frac{w}{w_a},
\end{aligned}
\label{equation for horizontal proposals}
\end{equation}
\begin{equation}
\begin{aligned}
{}&u_x^* = \frac{x^* - x_a}{w_a}, u_y^* = \frac{y^* - y_a}{h_a}, 
\\{}&u_h^* = \log\frac{h^*}{h_a}, u_w^* = \log\frac{w^*}{w_a},
\end{aligned}
\end{equation}
where \(x_a, x, x^*\) represent values related to anchors, the predicted boxes and the ground truth boxes, respectively, likewise for \(y, h, w\). We define transformation parameter regression target \(v^*\) as:
\begin{equation}
\begin{aligned}
{}&v_1^* = \frac{w_p}{w}\cos(\theta_p - \theta), v_2^* = -\frac{h_p}{h}\sin(\theta_p - \theta),
\\{}&v_{3}^* = \frac{w_p}{w}\sin(\theta_p - \theta), v_4^* = \frac{h_p}{h}\cos(\theta_p - \theta),
\end{aligned}
\end{equation}
where \(v_i^*(0\leq i < 4)\) are elements in the multiplication of rotation matrix and scaling matrix in Eq. \ref{rotation matrix} and Eq. \ref{scale matrix}. Specifically, \(\theta\) indicates the orientation of horizontal proposals, thus \(\theta = 0\). 

During training, we match H-anchors with the ground-truth bounding boxes based on IoUs between anchors and minimum horizontal enclosing rectangles of rotated ground truth. We assign a positive or negative label on anchors satisfying the conditions similar as RPN\cite{DBLP:conf/nips/RenHGS15}. The horizontal proposal (\(x, y, w, h, \theta\)) can be estimated from H-anchors with 4 variables (\(u_x, u_y, u_h, u_w\)) by Eq. \ref{equation for horizontal proposals}. Then we take the parameter tuple \(v\) as the input of Eq. \ref{equation for affine transformation} for obtaining oriented proposals. 

\begin{figure}[ht]
  \centering
  \includegraphics[width=7cm]{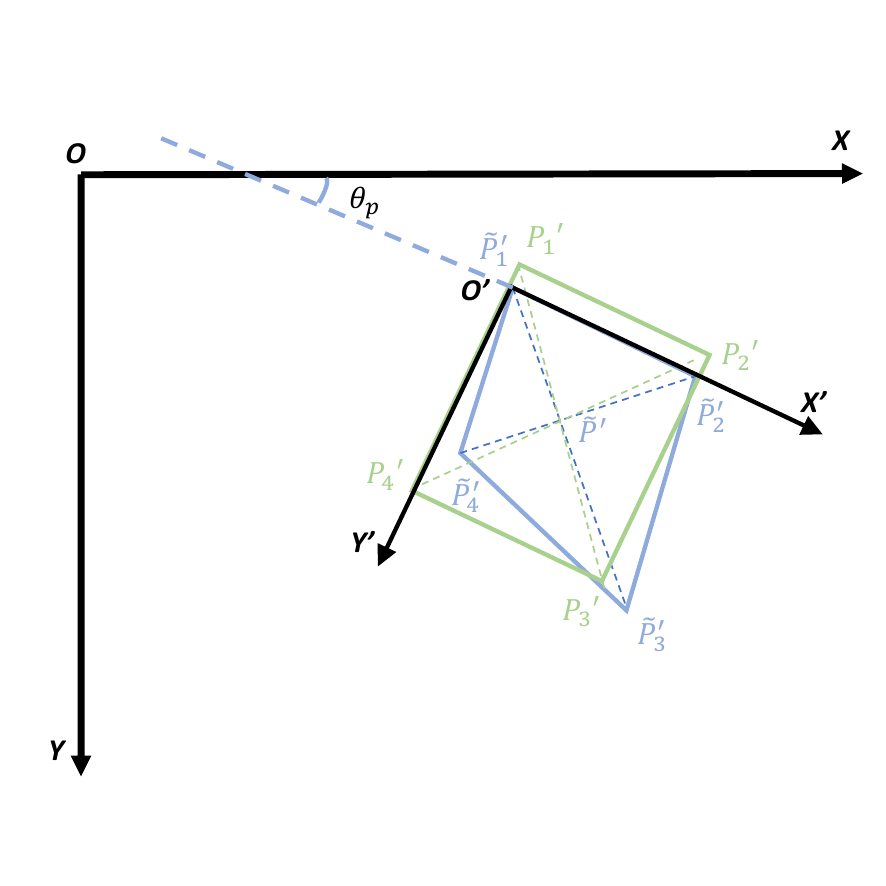}
  \caption{A quadrilateral predicted from AO-RPN and its corresponding rotated rectangle. The quadrilateral is represented by blue lines and the rotated rectangle is marked with green. The quadrilateral and rectangle share the same geometric center $\tilde{P}'$ and rotation angle $\theta_p$.}
  \label{Fig quad2rec}
\end{figure}

As discussed in \cite{DBLP:conf/eccv/0005Y20, DBLP:journals/corr/abs-1911-08299}, the predictions of our AO-RPN may be quadrilaterals, an example is depicted in Fig. \ref{Fig quad2rec}. We adjust the quadrilateral to form a rotated rectangle preparing for MH-Net according to the following steps. Firstly, we define the rotation angle $\theta_p$ of the blue quadrilateral as the angle between $x$-axis and $\tilde{P_1'}\tilde{P_2'}$ as mentioned above. For each quadrilateral, its corresponding adjoint rotated rectangle shares the same position $\tilde{P'} = (x_p, y_p)$ and the rotation angle $\theta_p$, as depicted in Fig. \ref{Fig quad2rec}. The width and height of the rotated rectangle are obtained as follows:
\begin{center}
\begin{equation}
\begin{aligned}
&w_p = \max \{|x_1^l - x_2^l|, |x_3^l - x_4^l|\},
\\&h_p = \max \{|y_1^l - y_4^l|, |y_2^l - y_3^l|\},
\end{aligned}
\end{equation}
\end{center}
where $(x_i^l, y_i^l) (i = 1, 2, 3, 4)$ are coordinates of $\tilde{P_i'}$ in the local coordinate system $X'O'Y'$ in Fig. \ref{Fig quad2rec}. Finally, we successfully transform a quadrilateral to its adjoint rotated rectangle $(x_p, y_p, w_p, h_p, \theta_p)$, which is more suitable for representation of a rotated object instance.

\subsection{RRoI Align}

After obtaining candidate regions from the RPN, subsequent procedures are to apply RoI Pooling\cite{DBLP:conf/iccv/Girshick15} or RoI Align\cite{DBLP:conf/iccv/HeGDG17} and extract features of the candidates. These two operations are typically applied to horizontal proposals, in this work, we adopt RRoI Align to extract features from the rotational proposals. 

Given an input feature map \(\mathcal{F} \in \mathbb{R}^{H \times W \times C}\) and a rotated proposal (\(x_p, y_p, w_p, h_p, \theta_p\)), RRoI Align outputs the proposal feature \(\mathcal{Y} \in \mathbb{R}^{K \times K \times C}\). The feature map is divided into \(K \times K\) bins whose sizes are \( \frac{h_p}{K} \times \frac{w_p}{K}\). For each \(bin_{(i,j)}\) \((0\leq i,j<K)\), the number of sampling points is set as \(k_s \times k_s\). We define the input feature as a global coordinate system and the feature in rotated proposals as a local coordinate system. Therefore, the local coordinates of sampling points in \(bin_{(i,j)}\) are in set \(\{i{h_p}/{K} + (i_h+0.5){h_p}/({Kk_s})|i_h=0,1,...,k_s-1\} \times\{j{w_p}/{K} + (j_w+0.5){w_p}/({Kk_s})|j_w=0,1,...,k_s-1\}\). We apply bilinear interpolation \(\mathcal{B}\) and average pooling in each bin as follows:
\begin{equation}
\mathcal{Y}(i, j) = 
\frac{1}{k_s \times k_s}\sum_{(x_l, y_l) \in bin_{(i,j)}} \mathcal{B}(\mathcal{F}, \mathcal{T}(x_l, y_l))
\end{equation}
where \((x_l, y_l)\) indicates the local coordinate in \(bin_{(i,j)}\) as mentioned above.
The function \(\mathcal{T}\) transforms a local coordinate to a global coordinate \((x_g, y_g)\) as:
\begin{equation}
\begin{pmatrix}x_g\\y_g\end{pmatrix} = 
\begin{pmatrix}\cos\theta_p & -\sin\theta_p\\ \sin\theta_p & \cos\theta_p\end{pmatrix}
\begin{pmatrix}x_l - w_p/2\\y_l - h_p/2\end{pmatrix} +
\begin{pmatrix}x_p\\y_p\end{pmatrix}
\end{equation}

During RRoI Align, we sample features inside rotated proposals and form horizontal feature maps with a fixed size of \(7 \times 7\). RRoI Align can sample rotated regions with arbitrary scales, aspect ratios and orientations.

\subsection{Multi-Head Network (MH-Net)}\label{Multi-Head Network}
Now, we have rotated proposals that potentially contain objects and features extracted by RRoI Align with almost clean background information. In this stage, Multi-Head Network predicts accurate bounding boxes with scores for the given categories. To overcome the misalignments between the classificaton and localization, we disentangle the detection task into multiple subtasks. And specifically, the location is divided into center point localization, scale prediction, and orientation estimation for providing more accurate bounding boxes regression. Each is achieved with elaborate designed architecture. In consequence, MH-Net has four sibling heads. And at the very end,  MH-Net integrates results from all branches into rotated bounding boxes \((x_r, y_r, w_r, h_r, \theta_r)\).

\begin{figure}[ht]
  \centering
  \includegraphics[width=9cm]{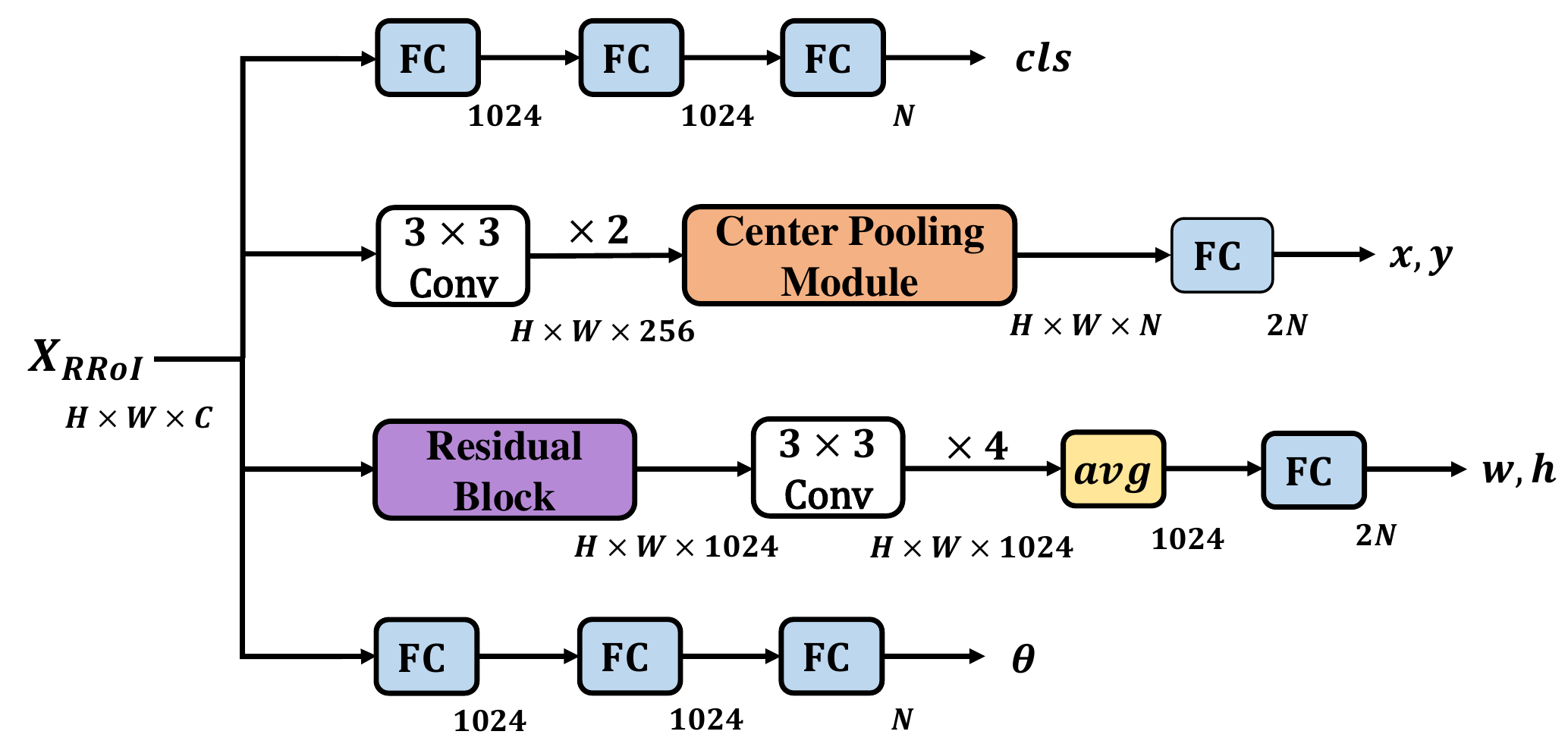}
  \caption{The structure of Multi-Head Network (MH-Net).}
  \label{Fig MH-Net}
\end{figure}

\begin{figure}[ht]
  \centering
  \includegraphics[width=5cm]{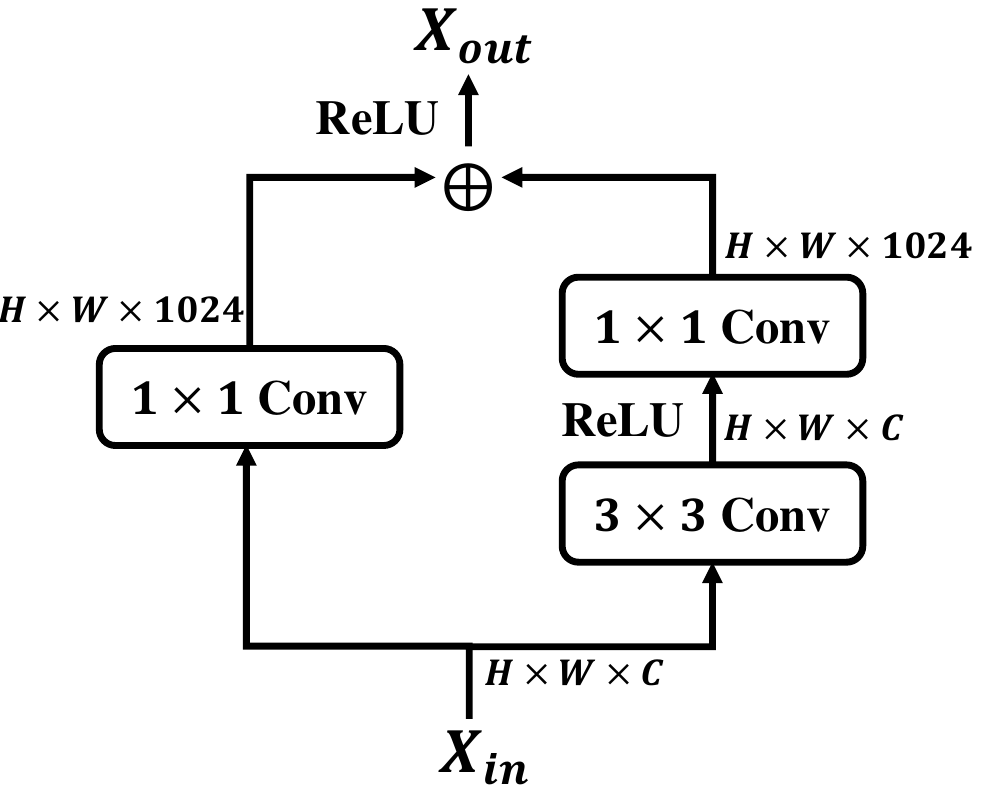}
  \caption{The residual block used in the scale head.}
  \label{res}
\end{figure}

\begin{figure}[ht]
  \centering
  \includegraphics[width=8cm]{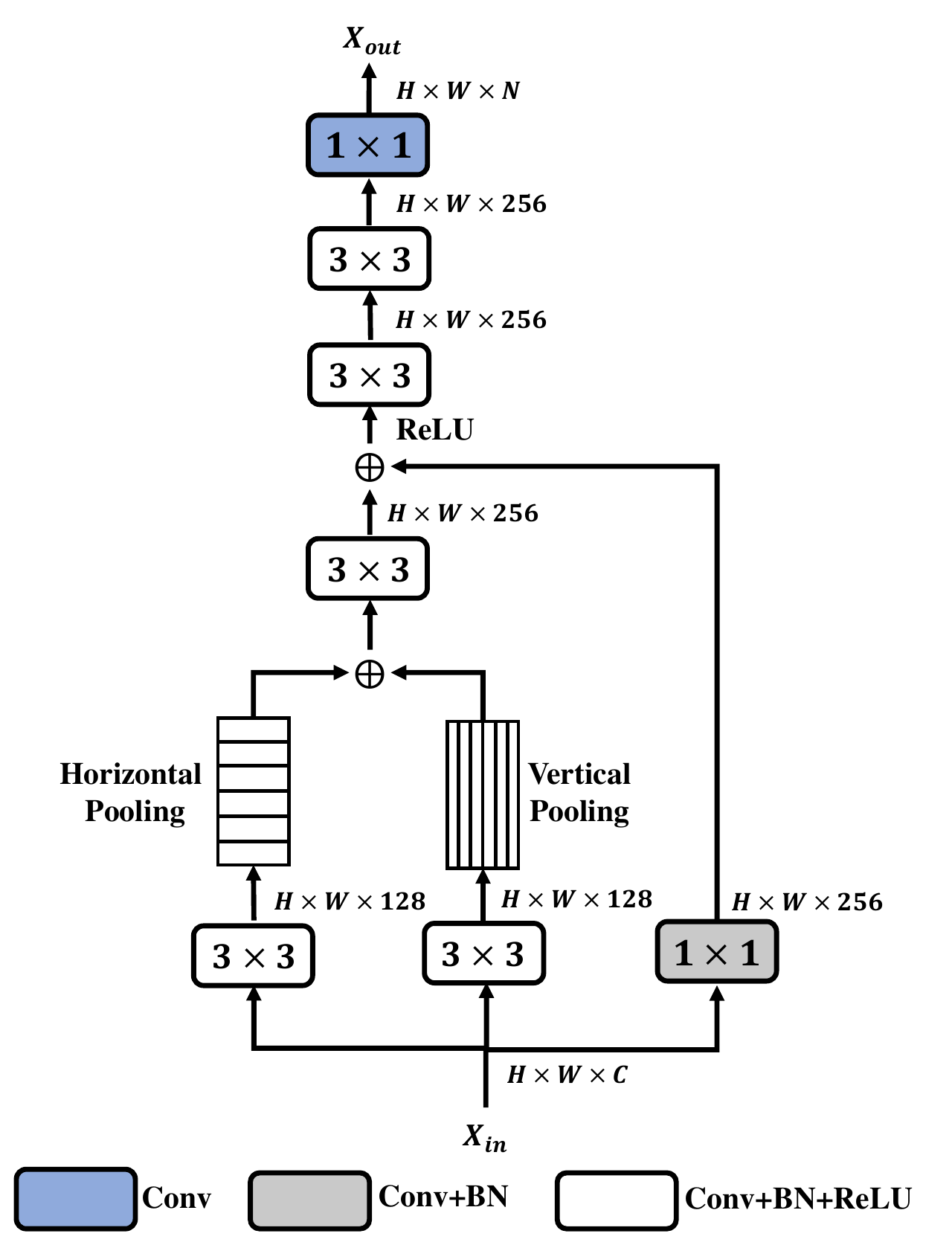}
  \caption{The architecture of the center pooling network.}
  \label{center pooling}
\end{figure}

As shown in Fig. \ref{Fig MH-Net}, MH-Net has a multi-branch structure with two fully connected branches and two convolutional branchs. The fully connected heads are for classification (\(cls\)) and orientation regression (\(\theta_r\)), respectively. Both of them have two 1024-dimension \textit{fc} layers, however, with different weights. The location head and scale head estimate the center coordinate \((x_r, y_r)\) and size \((w_r, h_r)\) of a rotated bounding box, respectively. The scale head is built with a residual module and four \(3\times 3\) convolution layers. The residual module is the same as that in Double Head RCNN\cite{wu2019double}, as shown in Fig. \ref{res}. It increases channels from \(C\) to \(1024\) for average pooling and 1024-d \(fc\) at the end of the scale branch. To improve location accuracy, we add a center pooling module\cite{DBLP:conf/iccv/DuanBXQH019} to the location head. The detailed architecture is shown in Fig. \ref{center pooling}. It returns a feature map with the size of \(H \times W \times N\), where \(N\) is the number of categories. Center pooling is composed of horizontal pooling and vertical pooling. In our implement, we maximize features in rows and columns simultaneously and sum the results together. For an RRoI, the output features in a channel indicates its geometric central information of one class, and determine if the location is a center point. After center pooling module, we use a \(fc\) layer to predict scale-invariant offsets in an RRoI's local coordinate system.

In the training stage, we calculate IoUs between rotated proposals and the ground truths using Eq. \ref{rotated iou} to determine positive and negative samplings,
\begin{equation}
IoU(R_i, G_j) = \frac{Area_{R_i\bigcap G_j}}{Area_{R_i\bigcup G_j}}
\label{rotated iou}
\end{equation}
where \(R_i\) and \(G_j\) represent rotated proposals and ground truths. \(Area\) means the geometric area of a rectangle. The assignment method is similar to AO-RPN as stated in Section \ref{Arbitrary-Oriented Region Proposal Network}. For MH-Net, proposals with IoUs greater than 0.5 are considered as positive samples. The predictions from the three sibling heads are defined as \(l\), \(s\), \(o\). \(l = (l_x, l_y)\), \(s = (s_h, s_w)\), and \(o = (o_{\theta})\). After RRoI Align, features are sampled from rotated candidates and transformed into horizontal feature maps with a fixed size of \(7 \times 7\). Then, we predict boxes in the RRoI's local coordinate system instead of a global coordinate system:
\begin{equation}
\begin{aligned}
& l_x = \frac{1}{w_p}((x_r - x_p)\cos\theta_p + (y_r - y_p)\sin\theta_p), 
\\ & l_y = \frac{1}{h_p}(-(x_r - x_p)\sin\theta_p + (y_r - y_p)\cos\theta_p),
\\ & s_h = \log\frac{h_r}{h_p}, \quad s_w = \log\frac{w_r}{w_p}, \quad o_{\theta} = \theta_r - \theta_p
\end{aligned}
\end{equation}
The regression targets for location \(l^*\), scale \(s^*\) and orientation \(o^*\) are calculated as follows: 
\begin{equation}
\begin{aligned}
& l_x^* = \frac{1}{w_p}((x_r^* - x_p)\cos\theta_p + (y_r^* - y_p)\sin\theta_p), 
\\ & l_y^* = \frac{1}{h_p}(-(x_r^* - x_p)\sin\theta_p + (y_r^* - y_p)\cos\theta_p),
\\ & s_h^* = \log\frac{h_r^*}{h_p}, \quad s_w^* = \log\frac{w_r^*}{w_p}, \quad o_{\theta}^* = \theta_r^* - \theta_p
\end{aligned}
\end{equation}
Here, \(x_p, x_r, x_r^*\) indicate the candidates, the predictions and the ground truths, likewise for \(y, w, h, \theta\). Note that we predict the minimum angle difference between an arbitrary-oriented candidate and its matched inclined ground truth, similar to AO-RPN. We use cross entropy loss function for classification and smooth L1 loss function for regression in three sibling heads. During testing, we apply NMS to suppress duplicated rotated bounding boxes and obtain the final results.

\section{Experiments}\label{experiments}

\begin{table*}[!t]
\renewcommand{\arraystretch}{1.3}
\caption{Ablation experiments of our method on DOTA. The abbreviation for categories are: PL-Plane, BD-Baseball diamond, BR-Bridge, GTF-Ground track field, SV-Small vehicle, LV-Large vehicle, SH-Ship, TC-Tennis court, BC-Basketball court, ST-Storage tank, SBF-Soccer ball field, RA-Roundabout, HA-Harbor, SP-Swimming pool, and HC-Helicopter. \(^*\) means adjusting orders of four points to regress minimal angles between horizontal proposals and oriented proposals. MH-the multi-head structure with four branches. CP-the center pooling module in the location branch. ANGLE-the structure of the orientation branch, e.g., fully connected head (fc) or convolutional head (conv).}
\label{Ablation experiments of our method on DOTA}
\centering

\resizebox{\textwidth}{!}{
  \begin{tabular}{ccccccccccccccccccccc}
  \hline
  
  \multirow{2}{*}{ }& \multirow{2}{*}{AO-RPN} & \multicolumn{3}{c}{MH-Net} &\multirow{2}{*}{PL}& \multirow{2}{*}{BD} & \multirow{2}{*}{BR} & \multirow{2}{*}{GTF} & \multirow{2}{*}{SV} & \multirow{2}{*}{LV} & \multirow{2}{*}{SH} & \multirow{2}{*}{TC} & \multirow{2}{*}{BC} & \multirow{2}{*}{ST} & \multirow{2}{*}{SBF} & \multirow{2}{*}{RA} & \multirow{2}{*}{HA} & \multirow{2}{*}{SP} & \multirow{2}{*}{HC} & \multirow{2}{*}{mAP} \\
   & & MH & CP & ANGLE & & & & & & & & & & & & & & & & \\
  \hline
  Baseline & & & & & 88.93 & 77.82 & 50.66 & 57.18 & 71.83 & 71.51 & 84.23 & 89.94 & 81.90 & 83.77 & 42.71 & 61.16 & 65.21 & 66.32 & 44.34 & 69.17\\
  Baseline\(^*\) & & & & & 88.73 & 83.26 & 53.09 & 54.80 & 76.52 & 74.47 & 86.00 & 90.56 & 85.22 & 83.53 & 49.62 & 63.08 & 72.56 & 69.04 & 65.43 & 73.06\\
  Baseline\(^*\) & $\surd$ & & & & 89.54 & 83.05 & 54.75 & 67.45 & 76.46 & 81.81 & 87.50 & 90.86 & 84.95 & 83.28 & 50.93 & 66.96 & 75.68 & 71.12 & 65.41 & 75.32\\
  Baseline\(^*\) & $\surd$ & $\surd$ & & conv & 89.45 & 83.93 & 55.14 & 67.12 & 77.06 & 83.04 & 87.53 & 90.89 & 86.74 & 83.92 & 53.31 & 66.92 & 76.20 & 71.02 & 65.86 & 75.87 \\
  Baseline\(^*\) & $\surd$ & $\surd$ & & fc & 89.57 & 82.56 & 54.64 & 67.63 & 76.76 & 82.74 & 87.71 & 90.85 & 87.82 & 84.42 & 50.75 & 68.69 & 76.83 & 71.39 & 64.69 & 75.74 \\
  Baseline\(^*\) & $\surd$ & $\surd$ & $\surd$ & conv & 89.51 & 82.84 & 54.60 & 67.23 & 76.59 & 82.58 & 87.70 & 90.88 & 83.75 & 84.58 & 53.96 & 65.80 & 75.95 & 71.20 & 63.13 & 75.35\\
  Baseline\(^*\) & $\surd$ & $\surd$ & $\surd$ & fc & 89.49 & 84.29 & 55.40 & 66.68 & 76.27 & 82.13 & 87.86 & 90.91 & 86.92 & 85.00 & 52.34 & 65.98 & 76.22 & 76.78 & 67.49 & 76.24\\
  \hline
  \end{tabular}}
\end{table*}

\begin{table*}[!t]
\renewcommand{\arraystretch}{1.3}
\caption{Performance comparison with others on DOTA(\%).}
\label{Performance comparison with others on DOTA.}
\centering
\resizebox{\textwidth}{!}{
\begin{tabular}{ccccccccccccccccccc}
\hline
Method & Backbone & FPN & PL & BD & BR & GTF & SV & LV & SH & TC & BC & ST & SBF & RA & HA & SP & HC & mAP \\
\hline
FR-O\cite{DBLP:conf/cvpr/XiaBDZBLDPZ18} & ResNet101 &  & 79.42 & 77.13 & 17.7 & 64.05 & 35.3 & 38.02 & 37.16 & 89.41 & 69.64 & 59.28 & 50.3 & 52.91 & 47.89 & 47.4 & 46.3 & 54.13\\
R-DFPN\cite{DBLP:journals/remotesensing/YangSFYSYG18} & ResNet101 & \checkmark & 80.92 & 65.82 & 33.77 & 58.94 & 55.77 & 50.94 & 54.78 & 90.33 & 66.34 & 68.66 & 48.73 & 51.76 & 55.10 & 51.32 & 35.88 & 57.94\\
ICN\cite{DBLP:conf/accv/AzimiVB0R18} & ResNet101 & \checkmark & 81.36 & 74.30 & 47.70 & 70.32 & 64.89 & 67.82 & 69.98 & 90.76 & 79.06 & 78.20 & 53.64 & 62.90 & 67.02 & 64.17 & 50.23 & 68.16\\
R\(^2\)CNN\cite{DBLP:journals/corr/JiangZWYLWFL17} & ResNet101 & & 80.94 & 65.67 & 35.34 & 67.44 & 59.92 & 50.91 & 55.81 & 90.67 & 66.92 & 72.39 & 55.06 & 52.23 & 55.14 & 53.35 & 48.22 & 60.67\\
RRPN\cite{DBLP:journals/tmm/MaSYWWZX18} & ResNet101 & & 88.52 & 71.20 & 31.66 & 59.30 & 51.85 & 56.19 & 57.25 & 90.81 & 72.84 & 67.38 & 56.69 & 52.84 & 53.08 & 51.94 & 53.58 & 61.01\\
RADet\cite{DBLP:journals/remotesensing/LiHPJS20} & ResNeXt101 & \checkmark & 79.45 & 76.99 & 48.05 & 65.83 & 65.46 & 74.40 & 68.86 & 89.70 & 78.14 & 74.97 & 49.92 & 64.63 & 66.14 & 71.58 & 62.16 & 69.09\\
RoI-Transformer\cite{DBLP:conf/cvpr/DingXLXL19} & ResNet101 & \checkmark & 88.64 & 78.52 & 43.44 & 75.92 & 68.81 & 73.68 & 83.59 & 90.74 & 77.27 & 81.46 & 58.39 & 53.54 & 62.83 & 58.93 & 47.67 & 69.56\\
CAD-Net\cite{DBLP:journals/tgrs/ZhangLZ19} & ResNet101 & \checkmark & 87.8 & 82.4 & 49.4 &  73.5 & 71.1 & 63.5 & 76.7 & 90.9 & 79.2 & 73.3 & 48.4 & 60.9 & 62.0 & 67.0 & 62.2 & 69.9\\
SCRDet\cite{DBLP:conf/iccv/YangYY0ZGSF19} & ResNet101 & \checkmark & 89.98 & 80.65 & 52.09 & 68.36 & 68.36 & 60.32 & 72.41 & 90.85 & \textbf{87.94} & \textbf{86.86} & 65.02 & 66.68 & 66.25 & 68.24 & 65.21 & 72.61\\
Gliding Vertex\cite{DBLP:journals/corr/abs-1911-09358} & ResNet101 & \checkmark & 89.64 & 85.00 & 52.26 & \textbf{77.34} & 73.01 & 73.14 & 86.82 & 90.74 & 79.02 & 86.81 & 59.55 & \textbf{70.91} & 72.94 & 70.86 & 57.32 & 75.02\\
Li et al.\cite{DBLP:conf/icip/LiXCWZY19} & ResNet101 & \checkmark & \textbf{90.21} & 79.58 & 45.49 & 76.41 & 73.18 & 68.27 & 79.56 & 90.83 & 83.40 & 84.68 & 53.40 & 65.42 & 74.17 & 69.69 & 64.86 & 73.28\\
Mask OBB\cite{DBLP:journals/remotesensing/WangDGCPY19} & ResNeXt-101 & \checkmark & 89.56 & \textbf{85.95} & 54.21 & 72.90 & 76.52 & 74.16 & 85.63 & 89.85 & 83.81 & 86.48 & 54.89 & 69.64 & 73.94 & 69.06 & 63.32 & 75.33\\
SARD\cite{DBLP:journals/access/WangZZZSG19} & ResNet101 & \checkmark & 89.93 & 84.11 & 54.19 & 72.04 & 68.41 & 61.18 & 66.00 & 90.82 & 87.79 & 86.59 & \textbf{65.65} & 64.04 & 66.68 & 68.84 & \textbf{68.03} & 72.95\\
FFA\cite{fu2020rotation} & ResNet101 & \checkmark & 90.1 & 82.7 & 54.2 & 75.2 & 71.0 & 79.9 & 83.5 & 90.7 & 83.9 & 84.6 & 61.2 & 68.0 & 70.7 & 76.0 & 63.7 & 75.7\\
ours & ResNet101 & & 89.41 & 83.28 & 51.63 & 69.32 & \textbf{76.94} & 74.06 & 79.00 & \textbf{90.87} & 80.06 & 83.00 & 46.92 & 67.48 & \textbf{76.55} & 70.19 & 65.63 & 73.62 \\
ours & ResNet101 & \checkmark & 89.49 & 84.29 & \textbf{55.40} & 66.68 & 76.27 & \textbf{82.13} & \textbf{87.86} & 90.81 & 86.92 & 85.00 & 52.34 & 65.98 & 76.22 & \textbf{76.78} & 67.49 & \textbf{76.24}\\
\hline
\end{tabular}}
\end{table*}

\begin{table*}[!t]
\renewcommand{\arraystretch}{1.3}
\caption{Performance Comparison with others on HRSC2016.}
\label{Performance Comparison on HRSC2016.}
\centering
\begin{tabular}{cccccccccc}
\hline
Method & CP\cite{DBLP:conf/icip/LiuHWY17} & BL2\cite{DBLP:conf/icip/LiuHWY17} & RC1\cite{DBLP:conf/icip/LiuHWY17} & RC2\cite{DBLP:conf/icip/LiuHWY17} & R\(^2\)PN\cite{DBLP:journals/lgrs/ZhangGZY18} & RRD\cite{DBLP:conf/cvpr/LiaoZSXB18} & RoI Transformer\cite{DBLP:conf/cvpr/DingXLXL19} & Gliding Vertex\cite{DBLP:journals/corr/abs-1911-09358} & ours\\
\hline
mAP(\%) & 55.7 & 69.6 & 75.7 & 75.7 & 79.6 & 84.3 & 86.2 & 88.20 & 89.94\\
\hline
\end{tabular}
\end{table*}

\subsection{Datasets}
We conduct extensive experiments on two popular and challenging benchmarks: DOTA\cite{DBLP:conf/cvpr/XiaBDZBLDPZ18} and HRSC2016\cite{DBLP:conf/icpram/LiuYWY17} to verify the effectiveness and superiority of our method. We report results with the standard protocol, i.e., Mean Average Precise (mAP).

\textbf{DOTA}\cite{DBLP:conf/cvpr/XiaBDZBLDPZ18} is a large-scale dataset for object detection in aerial images. It contains 2,806 aerial images collected from Google Earth, satellite JL-1, and so on. The image size in DOTA ranges from \(800 \times 800\) to \(4000 \times 4000\) pixels. There are a total of 188,282 instances with different scales, aspect ratios and orientations. Each instance is represented by a quadrilateral with four vertices \((x_1, y_1, x_2, y_2, x_3, y_3, x_4, y_4)\). 15 categories are included in the dataset: \textit{plane, ship, storage tank, baseball diamond, tennis court, swimming pool, ground track field, harbor, bridge, large vehicle, small vehicle, helicopter, roundabout, soccer ball field} and \textit{basketball court}. 
Many of them are densely distributed in the scenes. The maximum number of instances in one image is up to \(2000\), making it extremely challenging. In the experiments, we follow the standard protocol. \(1/2\) of images are selected as training set, \(1/6\) as validation set and \(1/3\) as testing set.

In the training stage, images with larger size are cropped into $1024 \times 1024$ patches with a stride of 824. If instances are divided into several parts, we discard them as adopted in \cite{DBLP:conf/cvpr/XiaBDZBLDPZ18}. 
Inference is also conducted on cropped images, we merge the results into the same resolutions with the original images. 

\textbf{HRSC2016}\cite{DBLP:conf/icpram/LiuYWY17} is a high resolution image dataset for ship detection. All the images are collected from six famous harbors. 
The resolutions of images in HRSC2016 range from \(300 \times 300\) to \(1500 \times 900\) pixels. There are 1,061 images in total, including 436 images for training, 181 images for validation and 444 images for testing. Ships in HRSC2016 are annotated by horizontal bounding boxes, oriented bounding boxes and pixel-wise segmentations. We use oriented bounding boxes for training and testing. In data pre-processing stage, we scale images to (512, 800) the same as \cite{DBLP:conf/cvpr/DingXLXL19}, where the length of the short side is 512 and the length of the long side is up to 800.

\subsection{Implementation Details}\label{implmentation details}
We build our model on top of FPN\cite{DBLP:conf/cvpr/LinDGHHB17} with ResNet101\cite{DBLP:conf/cvpr/HeZRS16} as backbone. We set anchor aspect ratios to \([0.5, 1, 2]\) for DOTA\cite{DBLP:conf/cvpr/XiaBDZBLDPZ18} and \([0.5, 1, 2, 1/3, 3]\) for HRSC2016\cite{DBLP:conf/icpram/LiuYWY17} due to large aspect ratios of ships. Same to FPN\cite{DBLP:conf/cvpr/LinDGHHB17}, \(\{P_2, P_3, P_4, P_5, P_6\}\) are built to generate anchors with different scales. Note that \(P_6\) is introduced only for a larger anchor scale. It is not used in the second stage of detection. Therefore, the total scales at each location are in set \(\{32, 64, 128, 256, 512\}\). During training AO-RPN, we choose \(256\) samples, where the number of positive and negative anchors are the same. Then we choose \(2000\) proposals which have overlaps between others lower than \(0.7\) by polygon NMS, likewise for testing. In the second stage, we randomly sample \(512\) proposals including \(128\) positive boxes for training. We retain bounding boxes with classification scores higher than \(0.05\) and set the IoU threshold of polygon NMS to \(0.1\) in the post-processing stage for testing.

The network is trained using Stochastic Gradient Descent (SGD) optimizer with momentum and weight decay setting to \(0.9\) and \(0.0001\) on 8 Geforce RTX 2080 Ti GPUs. We set mini batch size to 8, one for each GPU. Data augmentation including random horizontal flipping and random rotation with zero padding are adopted during training. The learning rate is initialized to \(0.01\) and divided by 10 at specific iterations. For DOTA\cite{DBLP:conf/cvpr/XiaBDZBLDPZ18}, we train the model for \(41k\) iterations with the learning rate decaying at \(\{27k, 37k\}\) iterations. For experiments on HRSC2016\cite{DBLP:conf/icpram/LiuYWY17}, the total training step is set to \(9.4k\) and the learning rate decays at \(\{6.2k, 8.6k\}\) steps.

\subsection{Ablation Study}
We conduct ablation experiments on DOTA\cite{DBLP:conf/cvpr/XiaBDZBLDPZ18} to validate the effectiveness of our network. We choose R\(^2\)CNN\cite{DBLP:journals/corr/JiangZWYLWFL17} based on Faster RCNN\cite{DBLP:conf/nips/RenHGS15} and FPN\cite{DBLP:conf/cvpr/LinDGHHB17} as baseline. The baseline has the same backbone and training and testing parameters with our method described in Section \ref{implmentation details}. We calculate mAP as a measure of performance.

\begin{figure}[ht]
\centering
\subfloat[]{
\centering
\includegraphics[width=4.1cm]{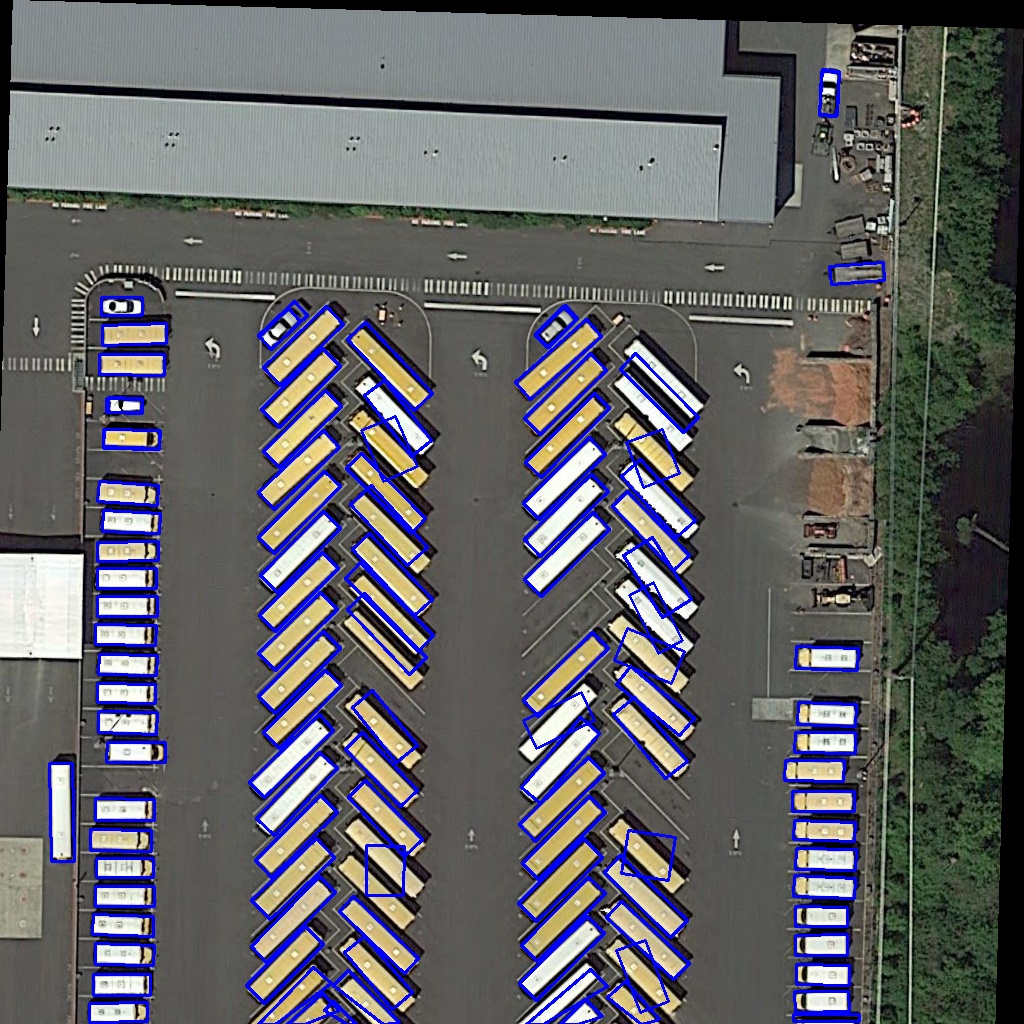}
\label{lv-example-baseline}}
\hfil
\subfloat[]{
\centering
\includegraphics[width=4.1cm]{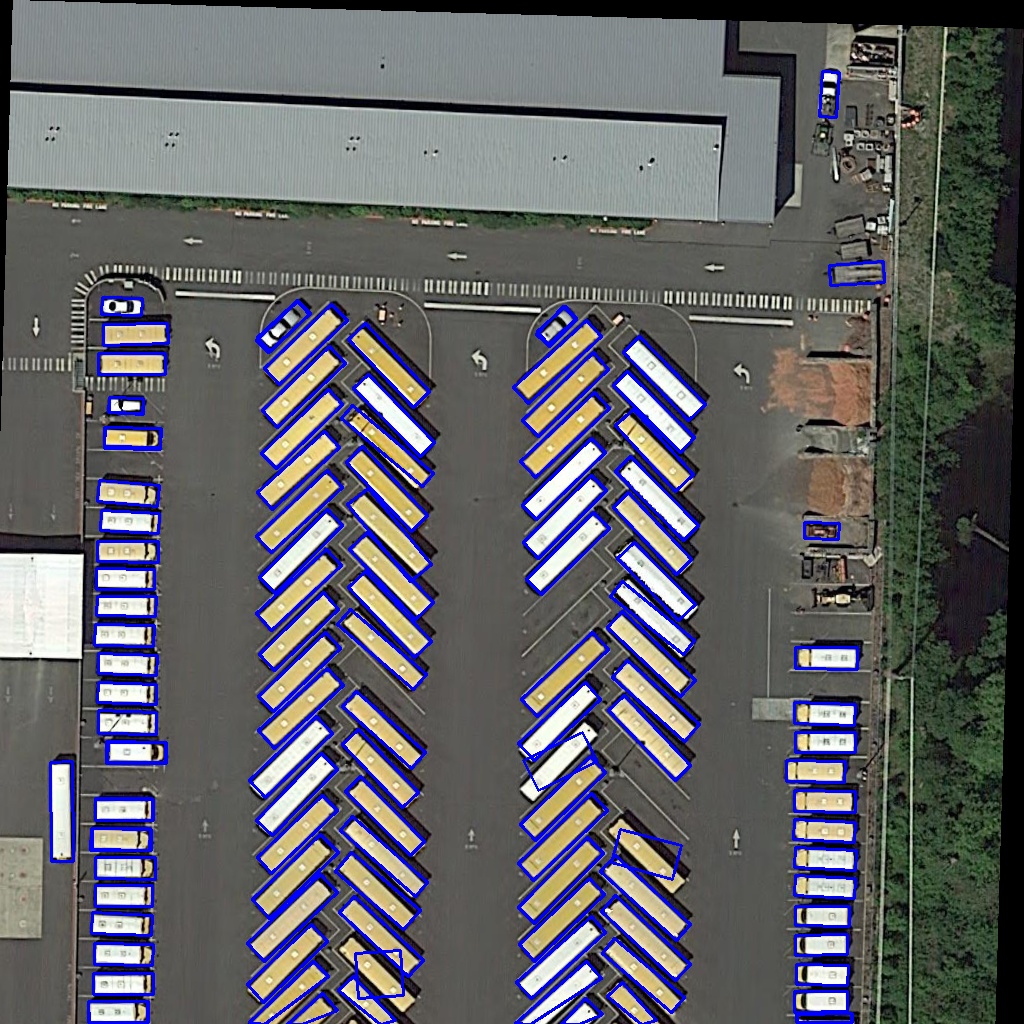}
\label{lv-example-aorpn}}
\caption{Examples of large vehicles on DOTA. (a) is the result of the baseline\(^*\). (b) is the result of baseline + AO-RPN.}
\label{lv-example}
\end{figure}

\begin{figure*}[!t]
\centering
\includegraphics[width=4.6cm,height=3cm]{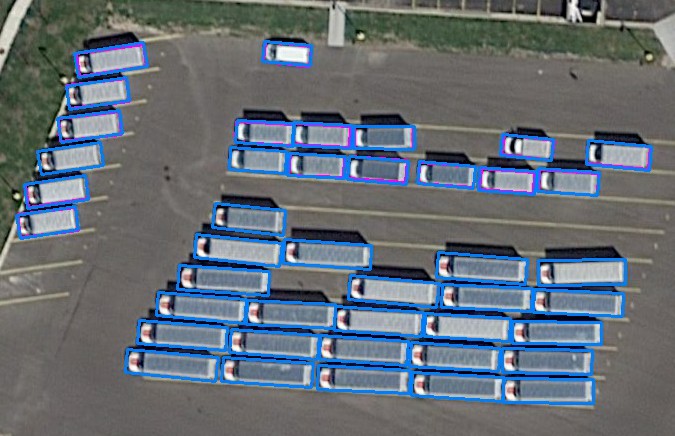}
\includegraphics[width=3.2cm,height=3cm]{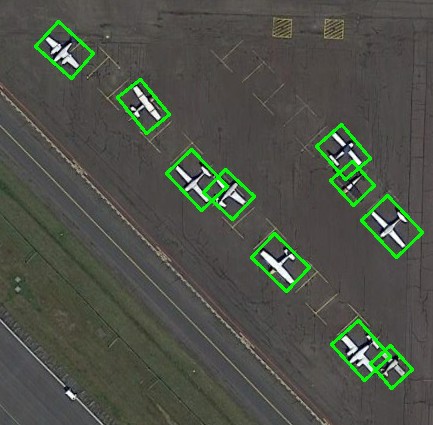}
\hfil
\includegraphics[width=3.2cm,height=3cm]{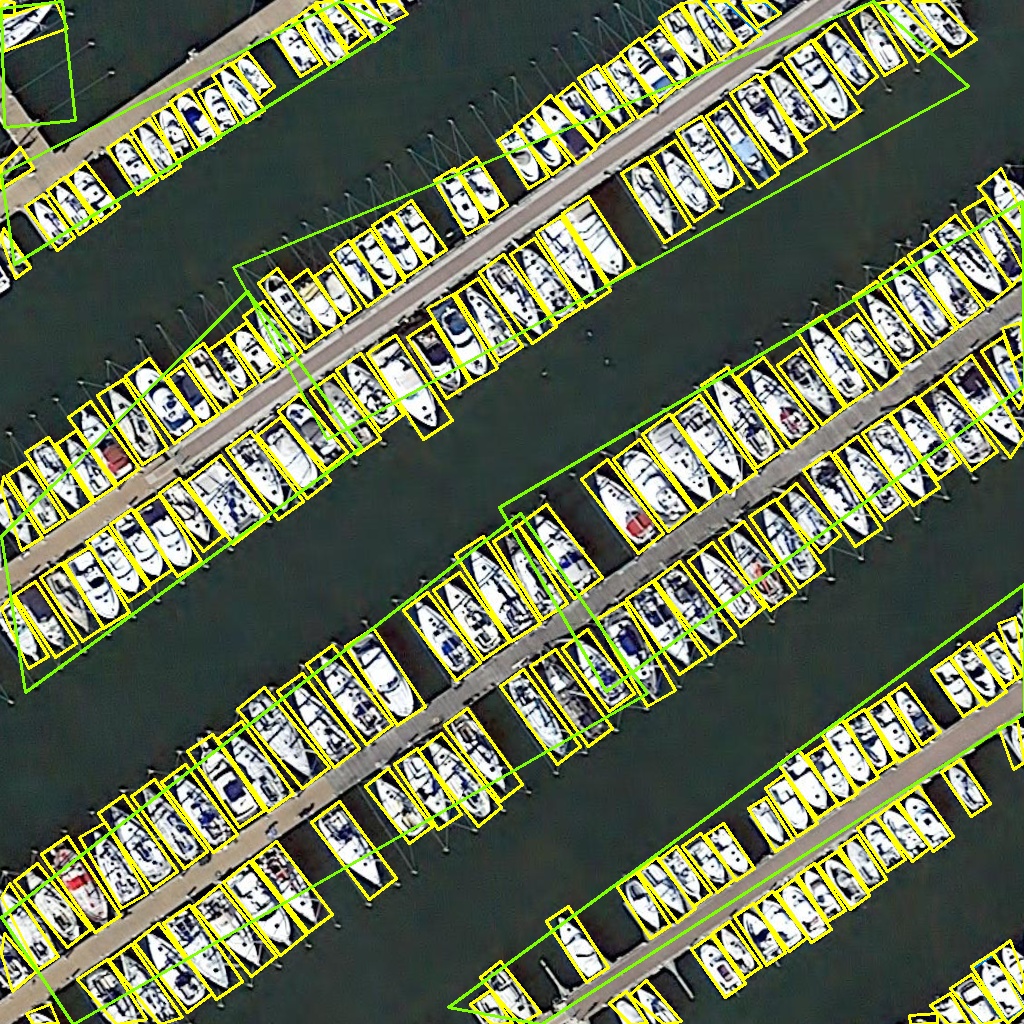}
\hfil
\includegraphics[width=3.2cm,height=3cm]{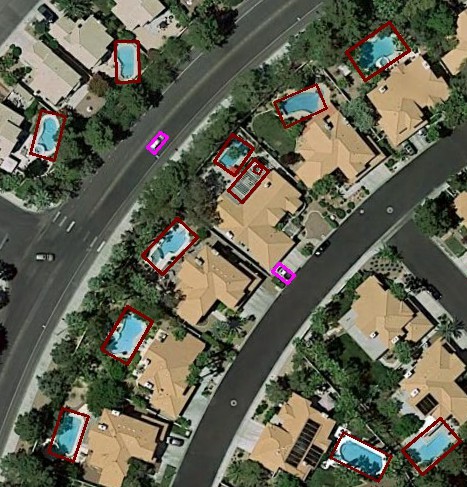}
\hfil
\includegraphics[width=3.2cm,height=3cm]{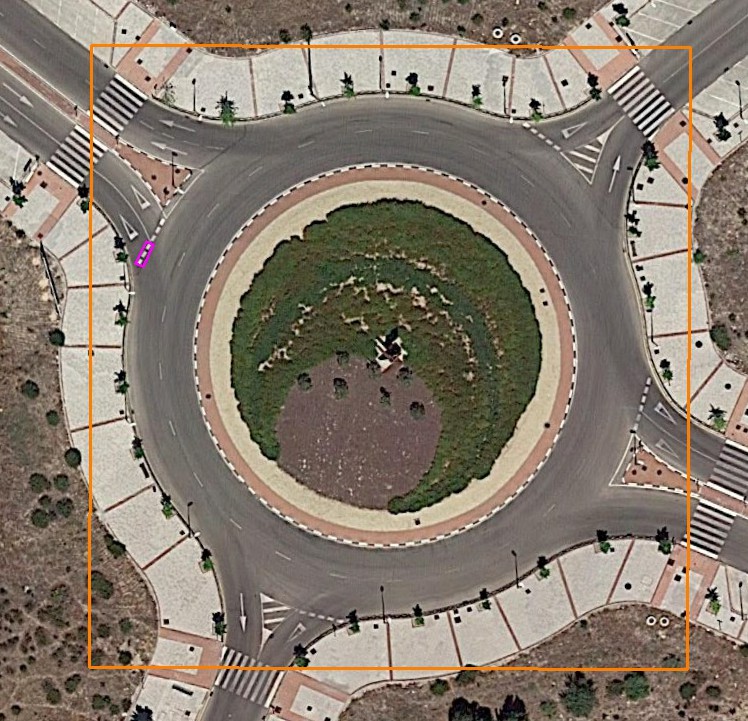}
\hfil
\vspace{2pt}
\includegraphics[width=3.1cm, height=3.2cm]{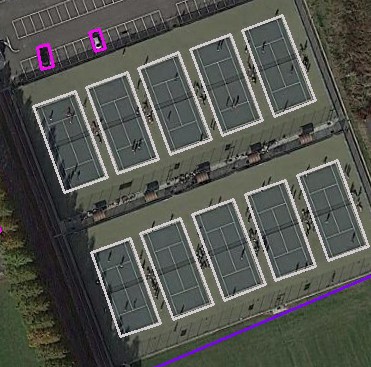}
\hfil
\includegraphics[width=3.2cm,height=3.2cm]{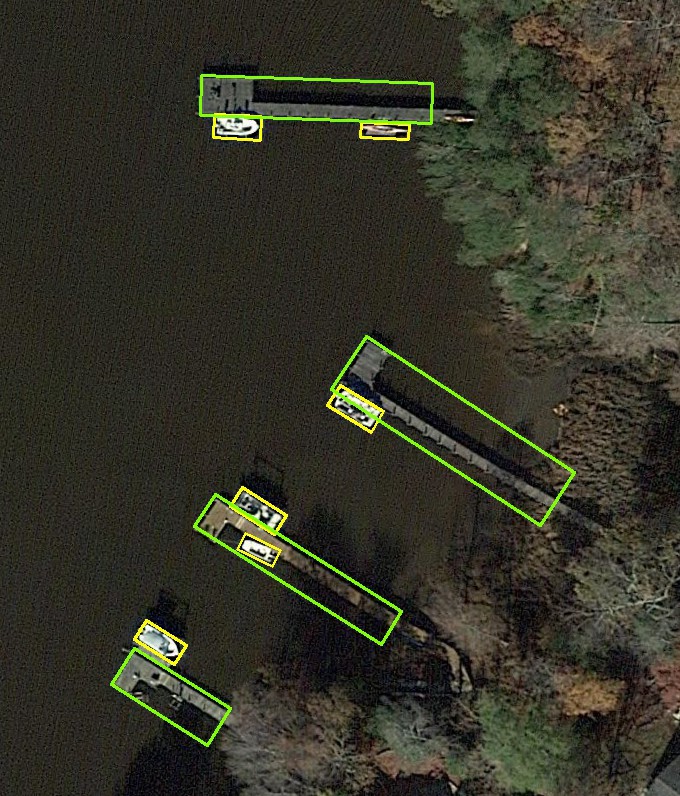}
\hfil
\includegraphics[width=3.3cm,height=3.2cm]{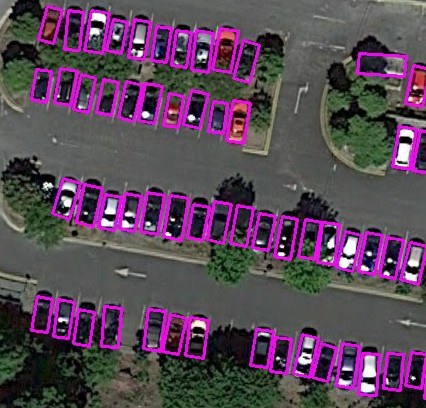}
\hfil
\includegraphics[height=3.2cm]{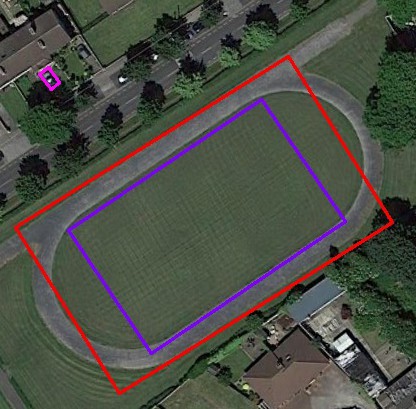}
\hfil
\includegraphics[width=4.6cm,height=3.2cm]{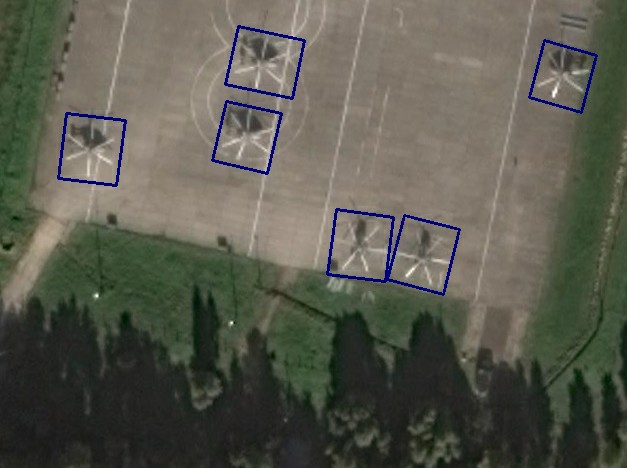}
\hfil
\vspace{2pt}
\includegraphics[width=5.0cm, height=3.15cm]{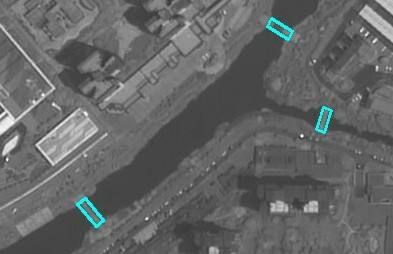}
\hfil
\includegraphics[width=3.3cm, height=3.15cm]{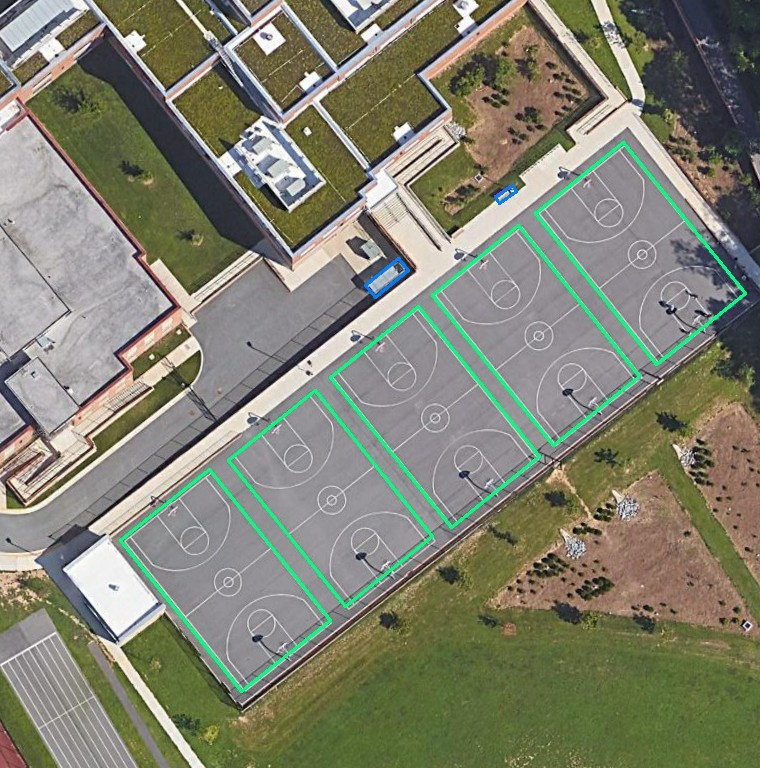}
\hfil
\includegraphics[width=4cm,height=3.15cm]{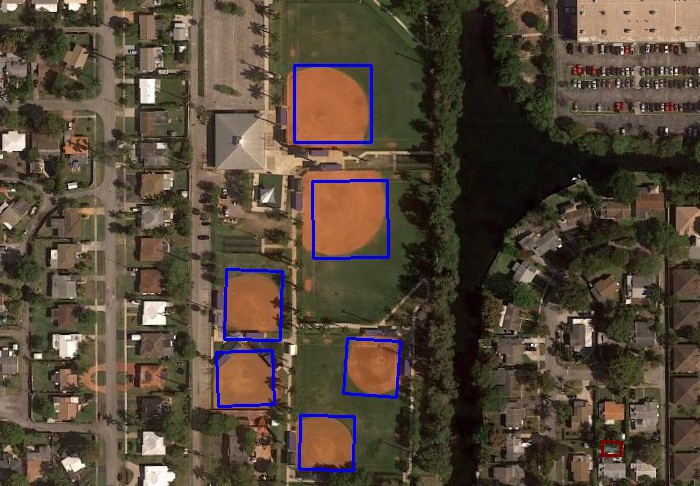}
\hfil
\includegraphics[width=5.0cm, height=3.15cm]{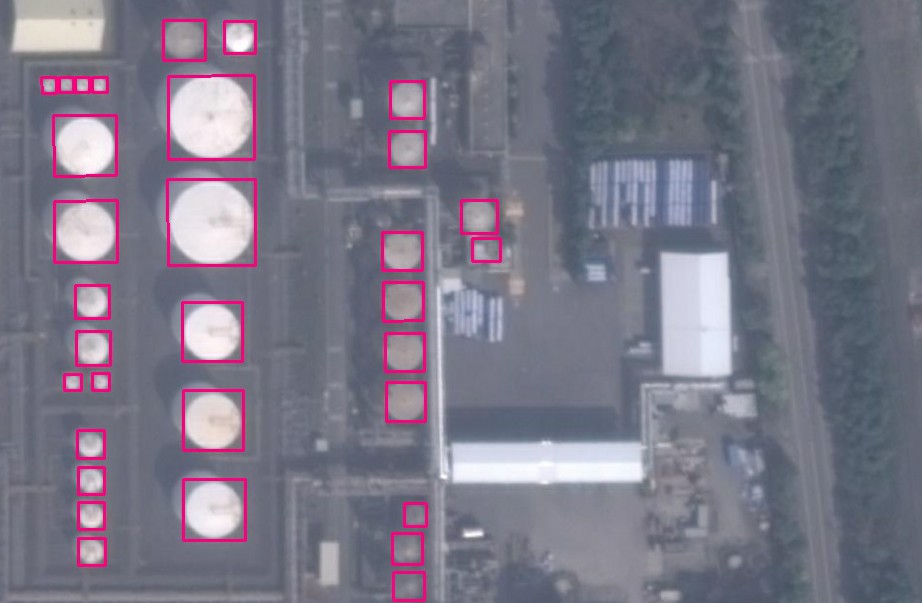}
\hfil
\includegraphics[width=18cm]{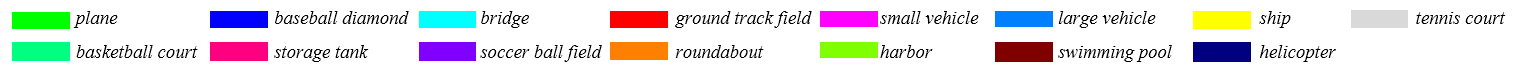}
\caption{Some detection results of our method on DOTA\cite{DBLP:conf/cvpr/XiaBDZBLDPZ18}.}
\label{dota-results}
\end{figure*}
    
\begin{figure*}[!t]
\centering
\includegraphics[width=3.5cm,height=2.5cm]{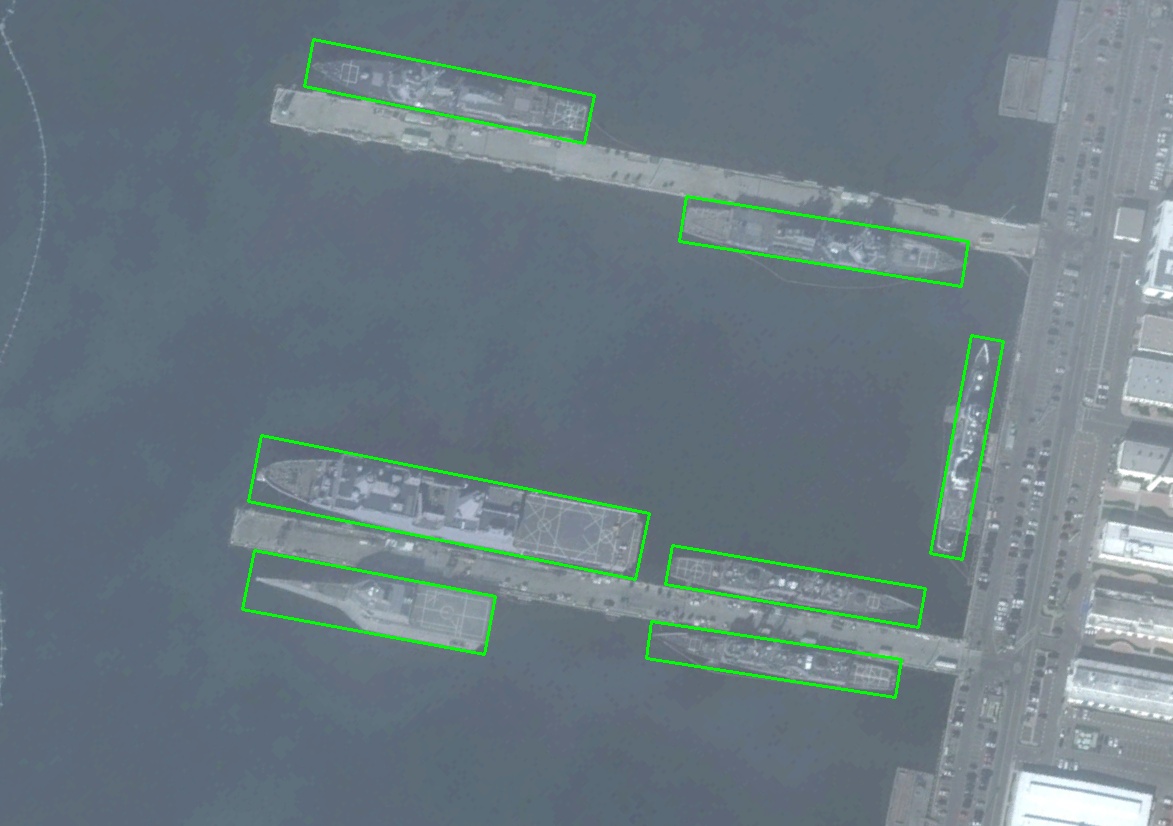}
\hfil
\includegraphics[width=3.5cm,height=2.5cm]{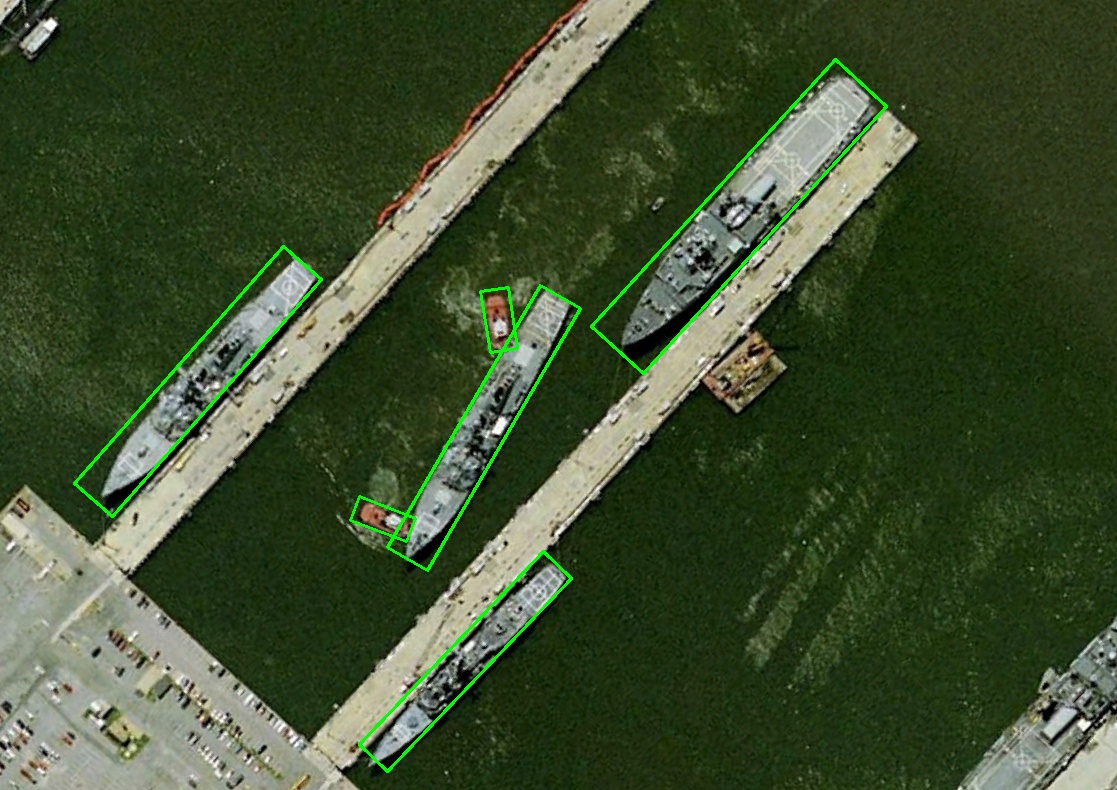}
\hfil
\includegraphics[width=3.5cm,height=2.5cm]{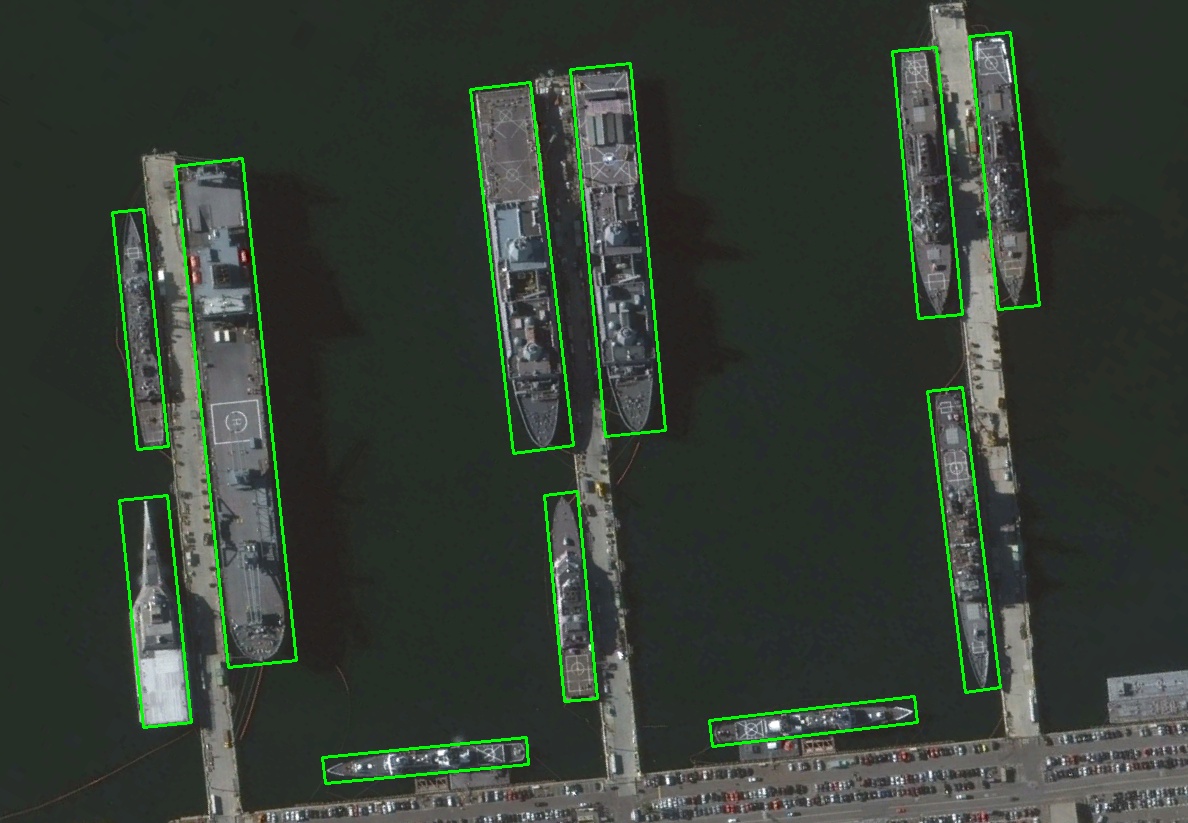}
\hfil
\includegraphics[width=3.5cm,height=2.5cm]{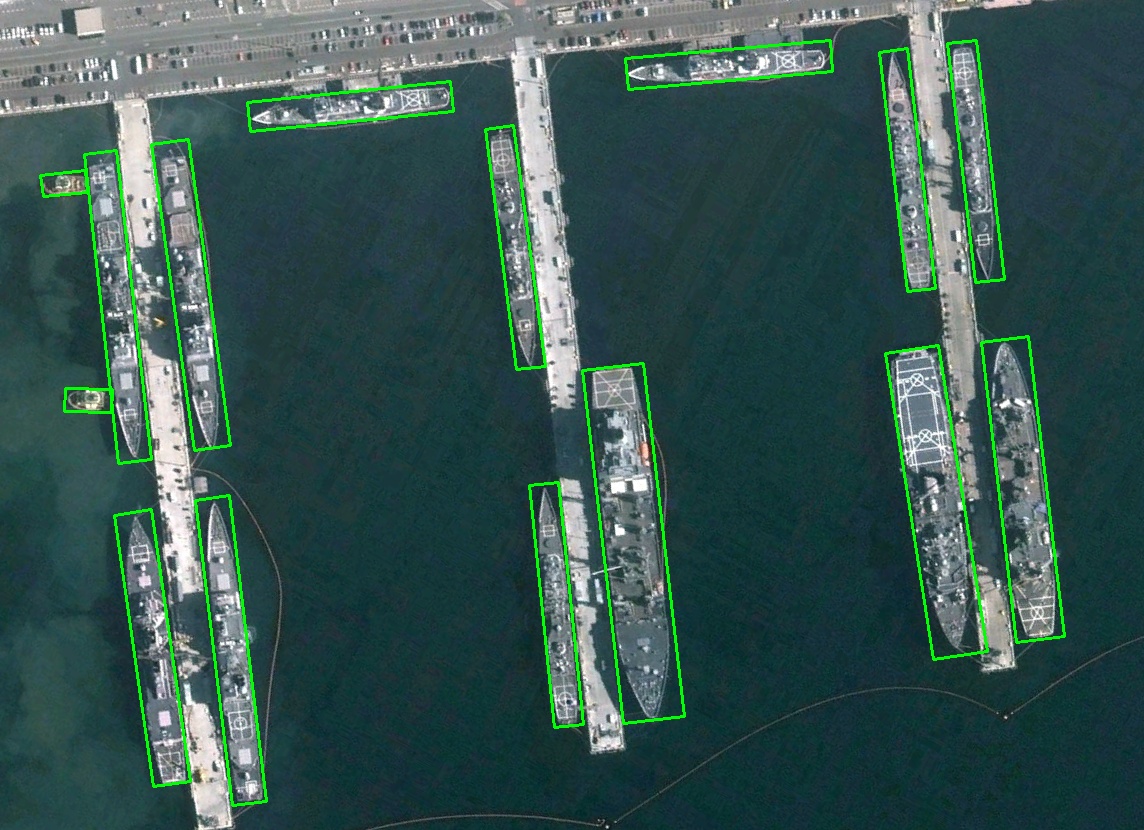}
\hfil
\includegraphics[width=3.5cm,height=2.5cm]{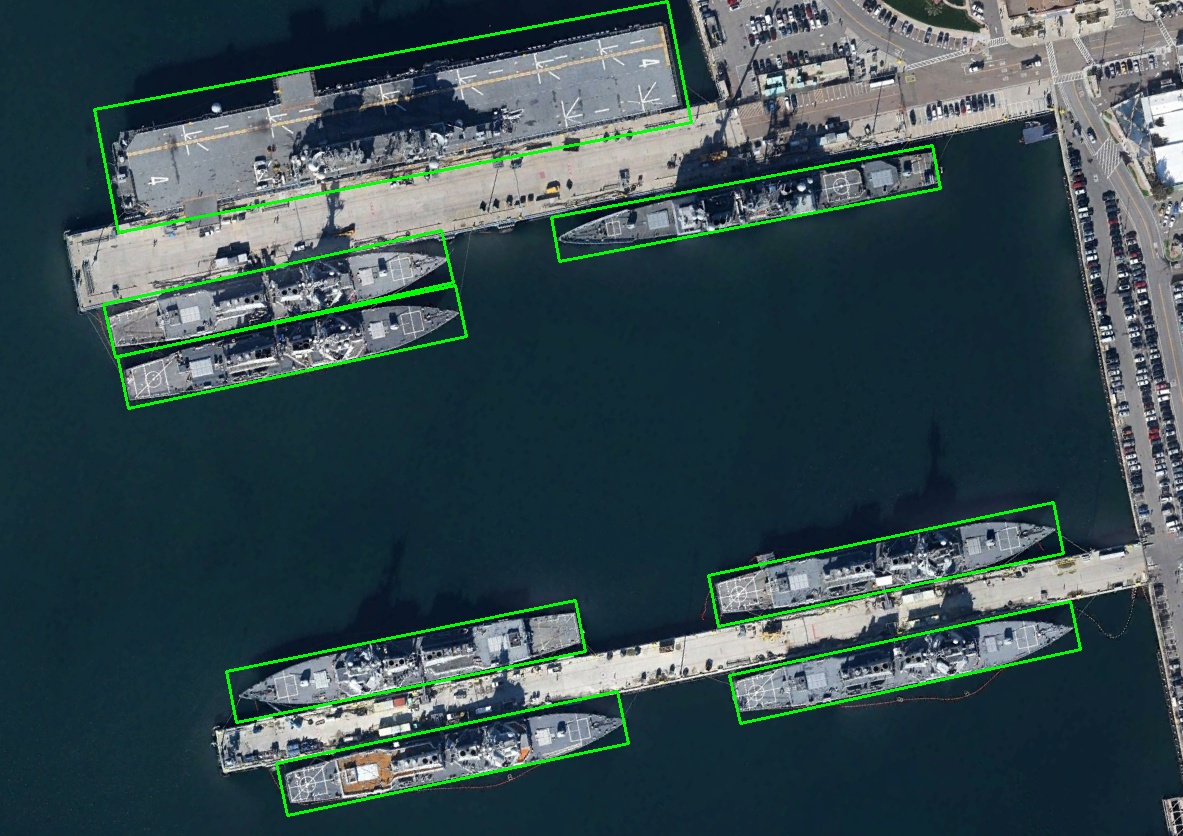}
\caption{Some detection results of our method on HRSC2016\cite{DBLP:conf/icpram/LiuYWY17}.}
\label{hrsc2016-results}
\end{figure*}

\subsubsection{AO-RPN}
The baseline defines angles in the way consistent with OpenCV, where orientation ranges in \([-\pi/2, 0)\) and set \(-\pi/2\) for horizontal proposals. Due to the point-order-based method used to define orientations in AO-RPN, we first change orders of four vertexes to minimize angles between horizontal proposals and their matched oriented proposals, and set orientation of horizontal boxes to 0 as described in Section \ref{Arbitrary-Oriented Region Proposal Network}. As shown in Table \ref{Ablation experiments of our method on DOTA}, there is approximately 3.89\% improvement in mAP. This is because adjusting orientation in advance restricts angles ranging from \(-\pi/4\) to \(\pi/4\). As a result, the amplitudes of angle offsets between horizontal proposals and their corresponding rotated proposals decrease from \(\pi/2\) to \(\pi/4\). And large regression targets for orientation are avoided. We then replace RPN and RoI Align in baseline with AO-RPN and RRoI Align, respectively. We find there is a 2.26\% improvement as shown in the third row of Table \ref{Ablation experiments of our method on DOTA}. The improvement is derived from two aspects. First, AO-RPN generates oriented proposals alleviating misalignments between proposals and objects. After RRoI Align, there is almost no background information sampled. Therefore, features inside objects account for a large proportion in local features of oriented proposals, which reduce the disturbance caused by noise outside objects. Second, different from the baseline predicting orientations only in the second stage, our method predicts oriented boxes twice in both region proposal stage and bounding box regression stage, which obtains more accurate scales and orientations. Moreover, we find there is an obvious increase in categories such as large vehicle by 7.04\% in mAP, which are usually densely distributed in regular scales and orientations. As shown in Fig. \ref{lv-example}, the baseline with AO-RPN performs better on dense oriented objects. The phenomenon indicates that RRoIs are appropriate for densely packed objects rather than HRoIs.

\begin{figure*}[!t]
  \centering
  \subfloat[{RoI Transformer\cite{DBLP:conf/cvpr/DingXLXL19}}]{
  \begin{minipage}[b]{0.23\linewidth}
  \includegraphics[width=4cm,height=4cm]{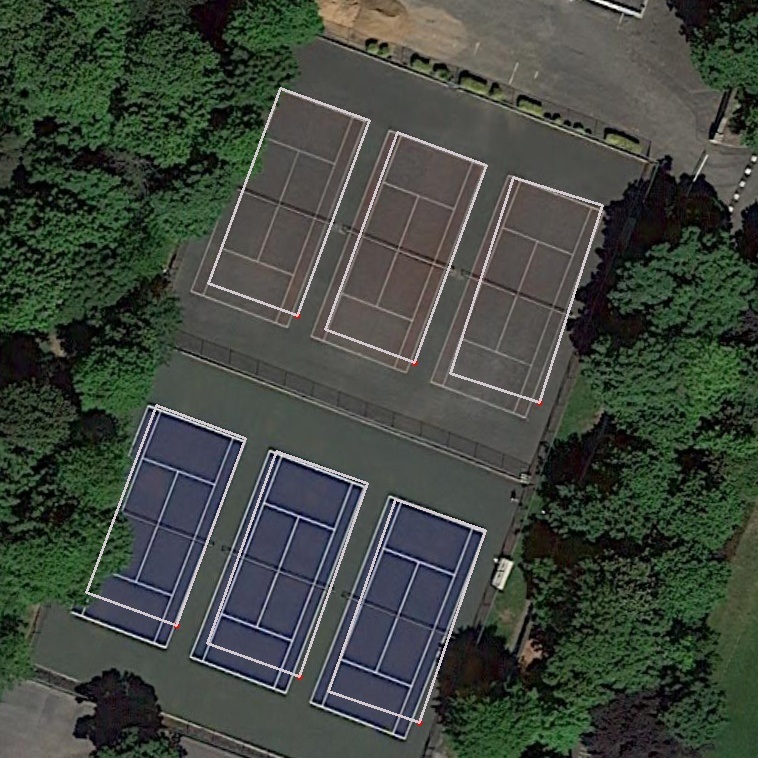}\vspace{4pt}
  \includegraphics[width=4cm,height=4cm]{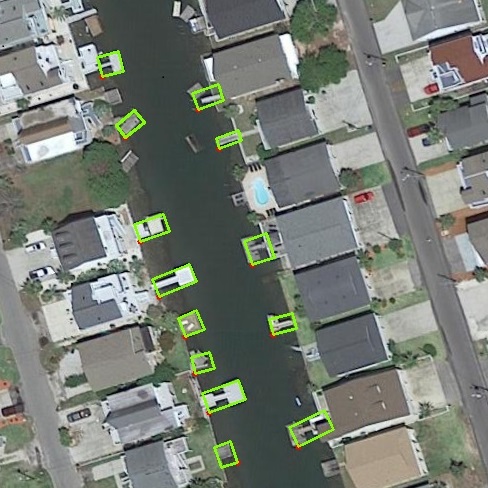}\vspace{4pt}
  \includegraphics[width=4cm,height=4cm]{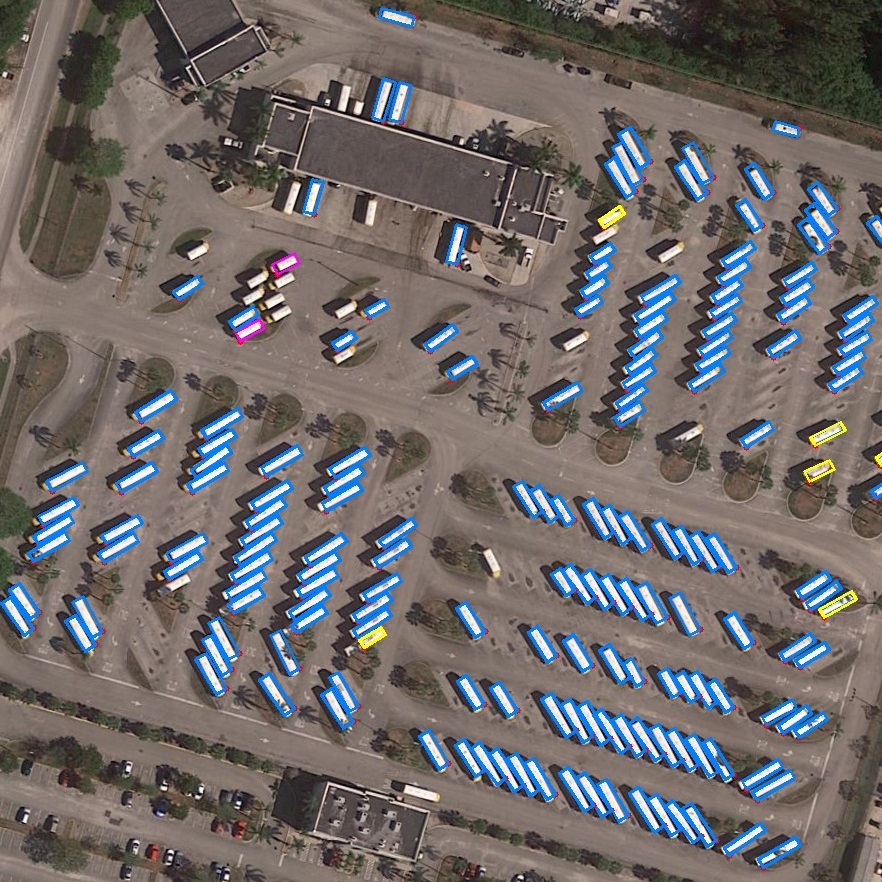}\vspace{4pt}
  \end{minipage}}
  \subfloat[{Gliding Vertex\cite{DBLP:journals/corr/abs-1911-09358}}]{
  \begin{minipage}[b]{0.23\linewidth}
  \includegraphics[width=4cm, height=4cm]{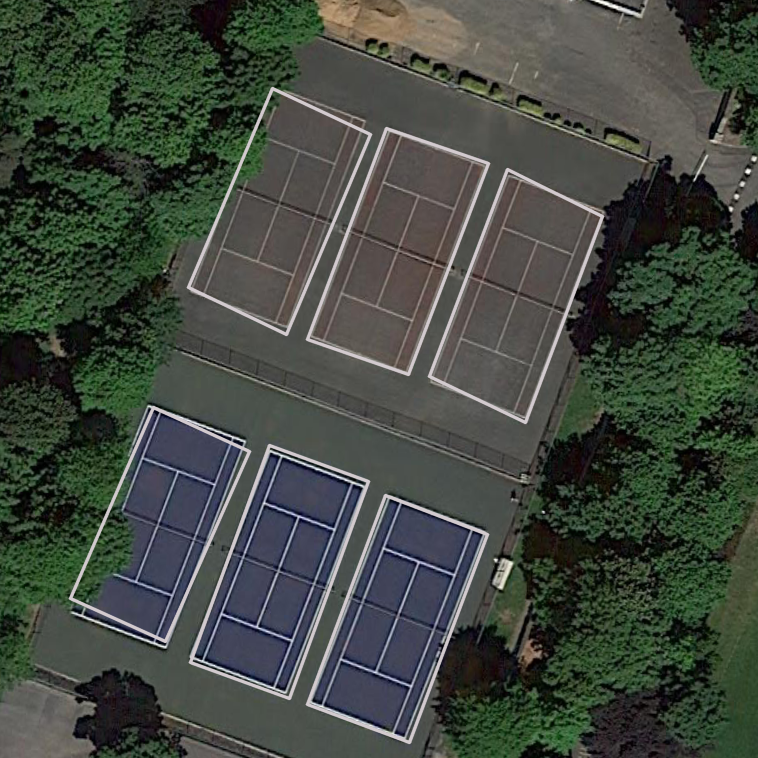}\vspace{4pt}
  \includegraphics[width=4cm,height=4cm]{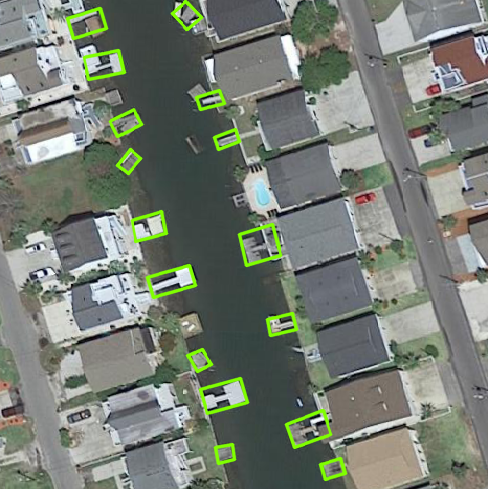}\vspace{4pt}
  \includegraphics[width=4cm,height=4cm]{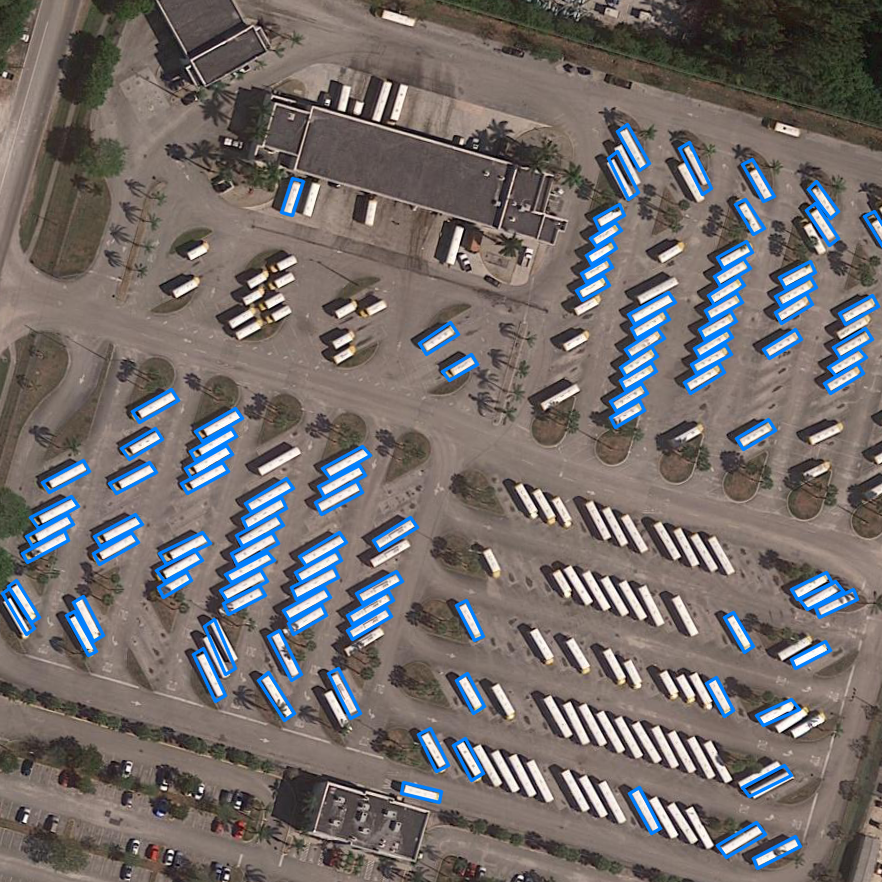}\vspace{4pt}
  \end{minipage}}
  \subfloat[SCRDet\cite{DBLP:conf/iccv/YangYY0ZGSF19}]{
  \begin{minipage}[b]{0.23\linewidth}
  \includegraphics[width=4cm,height=4cm]{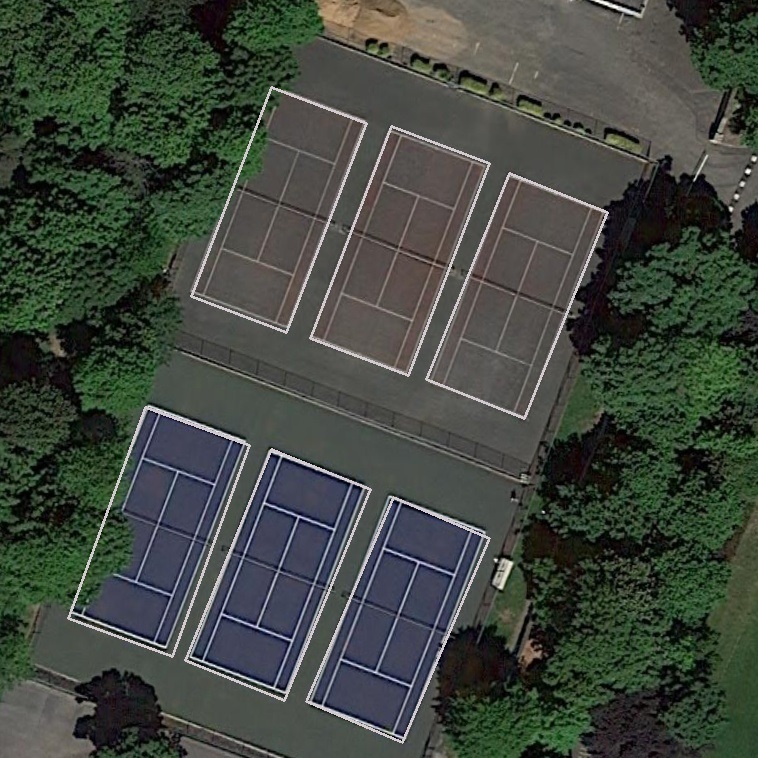}\vspace{4pt}
  \includegraphics[width=4cm,height=4cm]{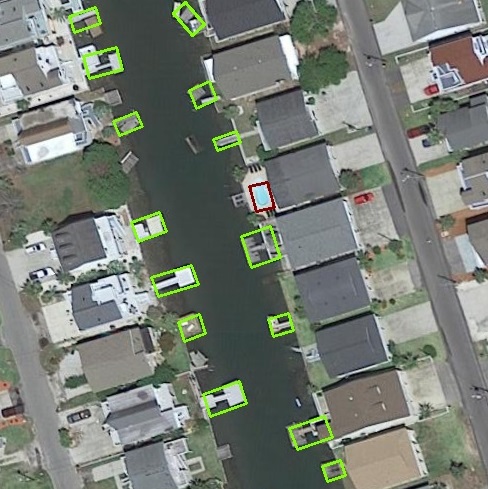}\vspace{4pt}
  \includegraphics[width=4cm,height=4cm]{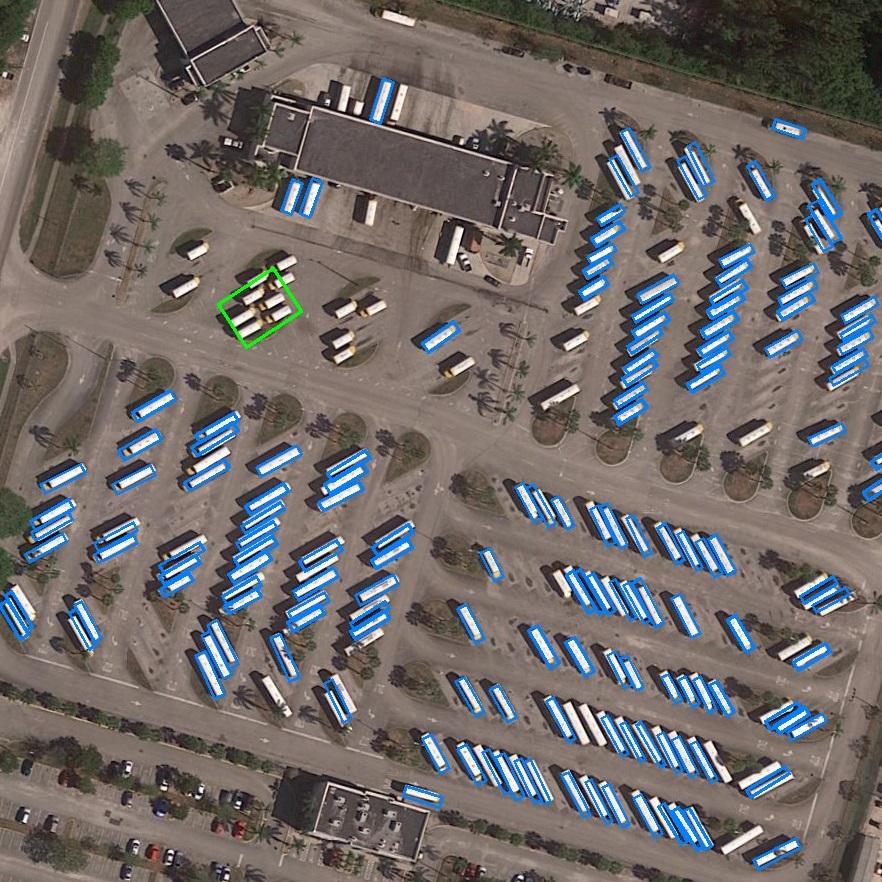}\vspace{4pt}
  \end{minipage}}
  \subfloat[Ours]{
  \begin{minipage}[b]{0.23\linewidth}
  \includegraphics[width=4cm,height=4cm]{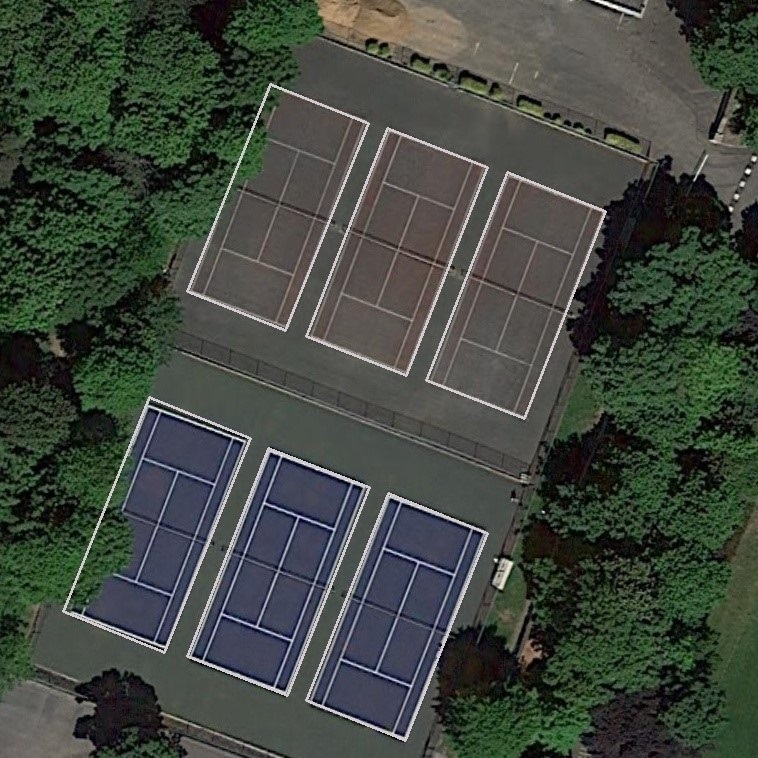}\vspace{4pt}
  \includegraphics[width=4cm,height=4cm]{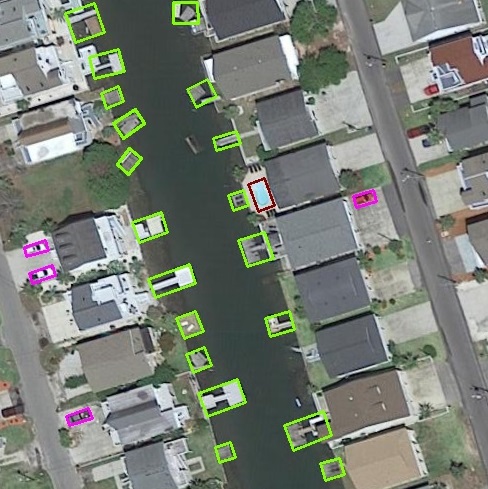}\vspace{4pt}
  \includegraphics[width=4cm,height=4cm]{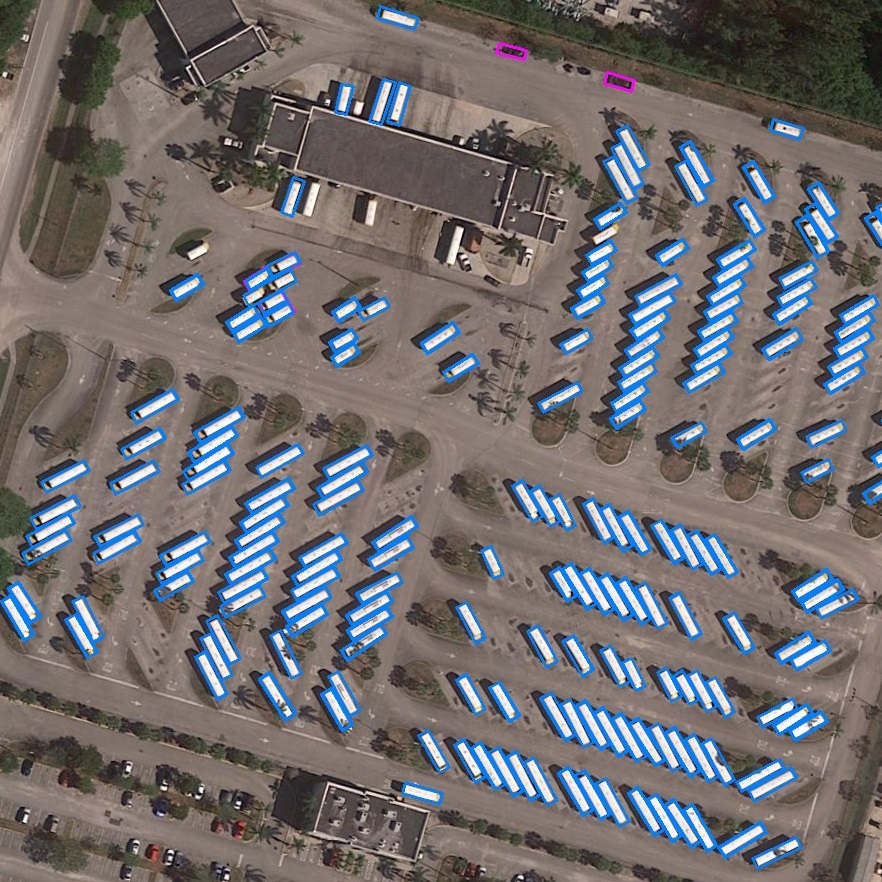}\vspace{4pt}
  \end{minipage}}
  \caption{Visual Comparison of detection results on DOTA\cite{DBLP:conf/cvpr/XiaBDZBLDPZ18}. Our method detects more accurate bounding boxes and misses fewer small objects}
  \label{comparison-dota}
  \end{figure*}

\subsubsection{MH-Net} 
The baseline uses a shared fully connected ($fc$s) head including \(2fc\)s to predict precise rotated bounding boxes in the second stage. We first replace the shared head with a multi-branch structure, which is similar to Fig. \ref{Fig MH-Net} where we use \(conv\)s instead of the center pooling network in the location branch and avoid the influence on feature extraction caused by reducing the convolution layers. We also conduct some comparative experiments with a convolutional head or a fully connected head for the orientation regression, which proves that $fc$ head is more suitable in our structure. From the fourth and fifth row in Table \ref{Ablation experiments of our method on DOTA}, we can see that the multi-branch structure without center pooling module gains 0.42\% and 0.55\% improvements in a fully connected head and a convolutional head for orientation prediction, respectively. The result indicates that it is reasonable to separate detection into four subtasks, i.e., classification, location, scale and orientation, and extract features in different branches. Moreover, after adding the center pooling network, the model with a convolutional orientation branch decreases 0.52\% in mAP and that with a fully connected orientation head improves detection results by about 0.5\%. The experiments show that the center pooling module could improve the localization performance with a fully connected orientation branch, which also demonstrate $fc$ is suitable for angle prediction in MH-Net.



\subsection{Comparisons with State-of-the-arts}     

In this section, we compare our method with the state-of-the-art methods on DOTA\cite{DBLP:conf/cvpr/XiaBDZBLDPZ18} and HRSC2016\cite{DBLP:conf/icpram/LiuYWY17}. The results are reported in Table \ref{Performance comparison with others on DOTA.} and Table \ref{Performance Comparison on HRSC2016.}.

\textbf{Results on DOTA}. 
MRDet is tested both with or without FPN. As shown in Table \ref{Performance comparison with others on DOTA.}, our method without FPN reaches a detection mAP of 73.62\%. It outperforms the previous methods without FPN (61.01\%) by 12.55 points and is better than some models with FPN, e.g., SCRDet\cite{DBLP:conf/cvpr/LiaoZSXB18} (72.61\%) and CAD-Net\cite{DBLP:journals/tgrs/ZhangLZ19} (69.9\%). There is a 2.62\% increase after we add FPN structure. Our method with FPN reaches the peak with an mAP of 76.24\% which is higher than the previous best result (FFA \cite{fu2020rotation}) by about 0.5\%. Moreover, MRDet achieves progress on some categories, such as \textit{large vehicle} and \textit{ship}, whose objects are often distributed in a high density with similar scales and orientations. For large vehicles and ships, MRDet achieves 82.13\% and 87.86\% in mAP with 2.23\% and 1.04\% improvements than the second best models (79.9\% and 86.82\%), respectively. 

We give some high quality visualized results on DOTA in Fig. \ref{dota-results}. From the cases, we can see our method performs well on aerial objects with arbitrary orientations, even in dense scenes. Compared with some state-of-the-art methods in Fig. \ref{comparison-dota}, e.g., RoI Transformer \cite{DBLP:conf/cvpr/DingXLXL19}, Gliding Vertex \cite{DBLP:journals/corr/abs-1911-09358} and SCRDet \cite{DBLP:conf/iccv/YangYY0ZGSF19}, our method regresses more tight and accurate bounding boxes and misses fewer small objects. In the second row in Fig.\ref{comparison-dota}, our model detects \textit{small vehicles} and \textit{swimming pools} in a large degree of perspective and low resolutions, while others mainly detect \textit{harbors} which occur frequently in the image. Besides, in the dense scene such as the third row in Fig. \ref{comparison-dota}, our method detects most of the objects and has better performance than others. However, MRDet fails to regress accurate bounding boxes on \textit{bridge} whose mAP is only 55.40\%. The performance of \textit{ground track field} also has a huge difference with the best result. We guess that may be due to large scales and aspect ratios of instances in these categories, which have huge differences with other categories' objects, e.g., the size of a bridge can be as large as 1200 pixels and about 98\% of the objects in DOTA are smaller than 300 pixels. This also may be resulted from the class imbalance under solved.

\textbf{Results on HRSC2016}.
HRSC2016 contains numerous long and narrow ships in large aspect ratios which are relatively fixed within a small range. Therefore, we add 1/3 and 3 in the aspect ratio set of the initialized anchors to detect slender objects. As shown in Tabel \ref{Performance Comparison on HRSC2016.}, our proposed method achieves 89.94\% in mAP, outperforming the second best method (88.20\%) by 1.74\%. Compared with RoI Transformer \cite{DBLP:conf/cvpr/DingXLXL19} whose anchor initialization scheme is similar to ours, the proposed method with a structurally simple network AO-RPN generating oriented proposals has a 3.74\% improvement. Some results on HRSC2016 are shown in Fig. \ref{hrsc2016-results}, which can be seen that our method detects high quality objects in different scales and orientations in spite of low luminosity and resolutions.

\section{Conclusion}\label{conclusion}
In this paper, we have presented an effective detector for oriented and densely packed objects in aerial images, called MRDet. We improve the region proposal stage by a lightweight network AO-RPN with the idea of adding a branch to learn affine transformation parameters from HRoIs to RRoIs. AO-RPN alleviates misalignments between proposals and objects efficiently without increasing the number of anchors and enhances the detection quality in high density scenes. Moreover, We predict classification confidences, locations, scales and orientations of the final bounding boxes by a multi-head network, which deals with features for different tasks by different structures. To locate objects more precisely, we innovatively add a center pooling module in the location branch. Our method does not only achieve state-of-the-art performance on popular datesets in aerial images, i.e., DOTA\cite{DBLP:conf/cvpr/XiaBDZBLDPZ18} and HRSC2016\cite{DBLP:conf/icpram/LiuYWY17}, but also outperforms on some categories with dense-packed and small-sized objects. However, the performance imbalance between different classes in multi-category datasets still exists. In the future, we would like to focus on the detection of those classes with low accuracy by analyzing characteristics of objects and applying better loss functions on the basis of focal loss\cite{DBLP:conf/iccv/LinGGHD17}.


%




\ifCLASSOPTIONcaptionsoff
  \newpage
\fi



%
\bibliographystyle{IEEEtran}
\bibliography{IEEEabrv, mybibfile}

\end{document}